\PassOptionsToPackage{unicode}{hyperref}
\PassOptionsToPackage{hyphens}{url}
\documentclass[
  11pt,
]{article}
\usepackage{amsmath,amssymb}
\usepackage{iftex}
\ifPDFTeX
  \usepackage[T1]{fontenc}
  \usepackage[utf8]{inputenc}
  \usepackage{textcomp} 
\else 
  \usepackage{unicode-math} 
  \defaultfontfeatures{Scale=MatchLowercase}
  \defaultfontfeatures[\rmfamily]{Ligatures=TeX,Scale=1}
\fi
\usepackage{lmodern}
\usepackage{newunicodechar}
\newunicodechar{−}{\ensuremath{-}}
\newunicodechar{Δ}{\ensuremath{\Delta}}
\newunicodechar{≈}{\ensuremath{\approx}}
\newunicodechar{≥}{\ensuremath{\geq}}
\newunicodechar{→}{\ensuremath{\rightarrow}}
\newunicodechar{×}{\ensuremath{\times}}
\newunicodechar{±}{\ensuremath{\pm}}
\newunicodechar{⁻}{\textsuperscript{-}}
\newunicodechar{⁷}{\textsuperscript{7}}
\newunicodechar{¹}{\textsuperscript{1}}
\newunicodechar{·}{\ensuremath{\cdot}}
\newunicodechar{³}{\textsuperscript{3}}
\newunicodechar{²}{\textsuperscript{2}}
\newunicodechar{⁶}{\textsuperscript{6}}
\newunicodechar{⁸}{\textsuperscript{8}}
\newunicodechar{ρ}{\ensuremath{\rho}}
\newunicodechar{ε}{\ensuremath{\varepsilon}}
\newunicodechar{τ}{\ensuremath{\tau}}
\newunicodechar{°}{\ensuremath{^\circ}}
\newunicodechar{≤}{\ensuremath{\leq}}
\newunicodechar{≠}{\ensuremath{\neq}}
\newunicodechar{∈}{\ensuremath{\in}}
\newunicodechar{↔}{\ensuremath{\leftrightarrow}}
\newunicodechar{‖}{\ensuremath{\|}}
\newunicodechar{√}{\ensuremath{\sqrt{}}}
\newunicodechar{⌊}{\ensuremath{\lfloor}}
\newunicodechar{⌋}{\ensuremath{\rfloor}}
\newunicodechar{λ}{\ensuremath{\lambda}}
\newunicodechar{α}{\ensuremath{\alpha}}
\newunicodechar{η}{\ensuremath{\eta}}
\newunicodechar{σ}{\ensuremath{\sigma}}
\newunicodechar{⊇}{\ensuremath{\supseteq}}
\newunicodechar{⁰}{\textsuperscript{0}}
\newunicodechar{⁴}{\textsuperscript{4}}
\newunicodechar{⁹}{\textsuperscript{9}}
\newunicodechar{…}{\dots}
\newunicodechar{¶}{\P}
\usepackage{float}
\usepackage{caption}
\ifPDFTeX\else
\fi
\IfFileExists{upquote.sty}{\usepackage{upquote}}{}
\IfFileExists{microtype.sty}{
  \usepackage[]{microtype}
  \UseMicrotypeSet[protrusion]{basicmath} 
}{}
\makeatletter
\@ifundefined{KOMAClassName}{
  \IfFileExists{parskip.sty}{%
    \usepackage{parskip}
  }{
    \setlength{\parindent}{0pt}
    \setlength{\parskip}{6pt plus 2pt minus 1pt}}
}{
  \KOMAoptions{parskip=half}}
\makeatother
\usepackage{xcolor}
\usepackage[margin=1in]{geometry}
\usepackage{longtable,booktabs,array}
\usepackage{calc} 
\usepackage{etoolbox}
\makeatletter
\patchcmd\longtable{\par}{\if@noskipsec\mbox{}\fi\par}{}{}
\makeatother
\IfFileExists{footnotehyper.sty}{\usepackage{footnotehyper}}{\usepackage{footnote}}
\makesavenoteenv{longtable}
\usepackage{graphicx}
\makeatletter
\def\maxwidth{\ifdim\Gin@nat@width>\linewidth\linewidth\else\Gin@nat@width\fi}
\def\maxheight{\ifdim\Gin@nat@height>\textheight\textheight\else\Gin@nat@height\fi}
\makeatother
\setkeys{Gin}{width=\maxwidth,height=\maxheight,keepaspectratio}
\makeatletter
\def\fps@figure{htbp}
\makeatother
\providecommand{\tightlist}{%
  \setlength{\itemsep}{0pt}\setlength{\parskip}{0pt}}
\ifLuaTeX
  \usepackage{selnolig}  
\fi
\usepackage{bookmark}
\IfFileExists{xurl.sty}{\usepackage{xurl}}{} 
\hypersetup{
  hidelinks,
  pdfcreator={LaTeX via pandoc}}

\author{}
\date{}

\begin{document}

\section{The Concept Allocation Zone}\label{the-concept-allocation-zone}

\textbf{Tracking How Concepts Form Across Transformer Depth}

\textbf{James Henry}\\
\emph{Independent Researcher} · jamesrahenry@henrynet.ca ·
\href{https://orcid.org/0009-0005-7126-9466}{ORCID 0009-0005-7126-9466}

Version 1: May 12, 2026 · \textbf{Version 2: August 12, 2026}

\emph{Version 2 corrects and retracts claims made in v1
(arXiv:2605.24856). Readers holding v1 should consult the ``Changes from
Version 1'' section at the end of this paper for the full list of
superseded values.}

\begin{center}\rule{0.5\linewidth}{0.5pt}\end{center}

\subsection{Abstract}\label{abstract}

Concept formation in transformer language models is a depth-extended
process, not a single-layer event: a concept becomes separable across
one or more contiguous regions of the residual stream --- its
\textbf{Concept Allocation Zone} (CAZ). A CAZ is not a concept but the
depth segment where the model organizes its geometry to make one
separable --- concepts may share a CAZ, and typically span multiple
across depth; the companion GEM paper {[}Henry, 2026b{]} shows the
separating direction continues to rotate within a CAZ before stabilizing
past its boundary. We formalize the CAZ through three layer-wise metrics
--- Separation, Concept Coherence, and Concept Velocity --- with
automated boundary detection that applies no significance threshold to
CAZ membership (every segment is a CAZ; ``strong'' vs.~``gentle'' is
score, never a binary cut). Empirical validation across 35 models, 8
architectural families, and 7 concepts shows the separation curve
\(S(l)\) is frequently \textbf{multimodal}, and scored detection
surfaces a further category of subtle allocation regions (``gentle
CAZes'') invisible to standard peak detection. The framework generates
seven testable predictions (§5); its contribution is the instrument and
the phenomena it surfaces --- the scored detector, the three metrics,
and the multimodal/gentle-CAZ findings --- not the predictions
themselves (full scorecard, §8). Released as the open-source
rosetta\_tools library (v1.3.1) {[}Henry, 2026, software{]}.

\begin{center}\rule{0.5\linewidth}{0.5pt}\end{center}

\subsection{1. Introduction}\label{introduction}

The dominant paradigm in mechanistic interpretability extracts concept
representations by identifying the single ``best layer'' --- the
residual stream depth at which a linear probe or difference-of-means
(DoM) vector achieves maximum class separation {[}Zou et al., 2023;
Arditi et al., 2024{]}. This heuristic is computationally convenient and
empirically grounded. It is also, by design, a snapshot: it identifies
the peak of a process rather than characterizing the process itself.

Transformers are iterative dynamical systems. Each layer applies a
sequence of attention and MLP operations that \emph{write} new
information into the residual stream, modifying and extending what prior
layers contributed {[}Elhage et al., 2021{]}. A concept observed at
Layer 15 was shaped by Layers 10 through 14 before it; the best-layer
heuristic tells us where the concept is most legible, not how it arrived
there.

This paper introduces the \textbf{Concept Allocation Zone} (CAZ)
framework, which extends the interpretability toolkit from anatomy ---
\emph{where is the concept most visible?} --- to dynamical flow ---
\emph{how does the concept form?} The CAZ is defined as a region of
model depth allocated to the geometric expression of a concept --- the
interval within which that concept becomes measurably separable. A CAZ
is not the concept itself --- each CAZ is a depth-extended segment of
the residual stream in which the model's geometry is organized to serve
a concept.

The framework has immediate practical implications:

\begin{enumerate}
\def\labelenumi{\arabic{enumi}.}
\tightlist
\item
  \textbf{Richer extraction.} CAZ/GEM (Geometric Evolution Maps; §4.2)
  diagnostics enable selecting a better single extraction layer (the
  handoff layer) than naive peak-layer probing (§4.4) --- the value is
  in \emph{which} layer CAZ/GEM identifies. Whether tracking depth this
  way outperforms a single-layer probe is a separate question the
  companion GEM paper addresses on its own terms {[}Henry, 2026b{]}.
\item
  \textbf{Intervention depth is encoding-strategy-dependent.} Ablation
  at different points in the CAZ chain produces qualitatively different
  effects in some models, but in models with redundant encoding,
  suppression is layer-invariant (§5.1, §5.8) --- so the framework
  characterizes \emph{when} layer choice matters rather than prescribing
  a single optimal intervention depth. The original prediction that
  mid-CAZ is optimal (P1) is not supported (§5.1).
\item
  \textbf{Dark matter connection.} The \emph{nonlinear} component of
  sparse-autoencoder (SAE) reconstruction error --- the part Engels et
  al.~{[}2025a{]} find is not linearly predictable from the input
  activation, a minority of the error norm --- may partially correspond
  to in-progress concept construction within CAZes: transitional
  representations that resist linear decomposition at any single layer.
  This is a testable hypothesis; direct cross-validation of CAZ regions
  against SAE residuals has not been conducted in this work.
\item
  \textbf{Cross-model transfer.} Concept directions may be
  depth-matchable across architectures --- a potential geometric basis
  for transferable interpretability tooling. Whether a single Procrustes
  rotation suffices to align all concepts simultaneously, and whether
  cross-architecture alignment is genuinely depth-matched, is the
  subject of dedicated follow-up work outside this paper's scope.
\item
  \textbf{Understanding alignment training.} CAZ profiles provide a lens
  for studying what alignment training changes in a base model's concept
  geometry --- not whether concepts exist, but where and how they are
  allocated. In principle CAZ profiles support a metric for quantifying
  alignment-training distortion --- the change in concept-selective
  feature count and alignment strength from base to instruct variant ---
  which is left to follow-up work.
\item
  \textbf{Concept inventory.} By tracking which geometric directions are
  allocated at each CAZ and which remain unaligned with any human
  concept probe, the framework provides a systematic approach to
  cataloging what a model computes --- both the named and the unnamed.
\end{enumerate}

The paper's positive contribution is first-party: the scored detector
and its three metrics, and the multimodal and gentle-CAZ phenomena they
surface (§4--§6). The framework also generates seven specific,
falsifiable predictions, fixed before this analysis began; §5 reports on
all seven, giving the per-prediction verdicts and the model subset each
was evaluated on, summarized in §8. Where they missed, following up on
why produced some of the framework's most useful results --- the
two-regime encoding structure (§5.8), the gentle-CAZ category, and the
handoff-layer question that grew into the companion GEM paper {[}Henry,
2026b{]}. The reference implementation is provided as rosetta\_tools
{[}Henry, 2026, software{]}, an open-source Python library. We are
explicit about the assumptions the framework inherits from the broader
interpretability literature.

\textbf{Corpus and analysis coverage.} Two companion papers are cited
throughout: the \textbf{GEM paper} {[}Henry, 2026b{]} (handoff-layer
extraction) and the \textbf{validation paper} {[}Henry, 2026c{]}
(large-scale causal ablation). Each uses a model set scoped to its
analysis; the table below records the scope rationale for each:

\begin{longtable}[]{@{}
  >{\raggedright\arraybackslash}p{(\columnwidth - 4\tabcolsep) * \real{0.3333}}
  >{\raggedright\arraybackslash}p{(\columnwidth - 4\tabcolsep) * \real{0.3333}}
  >{\raggedright\arraybackslash}p{(\columnwidth - 4\tabcolsep) * \real{0.3333}}@{}}
\toprule\noalign{}
\begin{minipage}[b]{\linewidth}\raggedright
Analysis
\end{minipage} & \begin{minipage}[b]{\linewidth}\raggedright
Coverage
\end{minipage} & \begin{minipage}[b]{\linewidth}\raggedright
Scope rationale
\end{minipage} \\
\midrule\noalign{}
\endhead
\bottomrule\noalign{}
\endlastfoot
CAZ detection and predictions (this paper) & 35 models, 7 concepts & 29
base + 6 instruct; full architectural range, retained from the initial
release \\
Direction-specificity ablation (this paper) & 16 of 35 models & Extended
to all 28 base models in {[}Henry, 2026c{]} \\
GEM handoff validation {[}Henry, 2026b{]} & 29 base models, 17 concepts
& Base models only; concept set expanded for handoff analysis \\
Full ablation sweep {[}Henry, 2026c{]} & 28 base models, 17 concepts &
Instruct variants excluded; systematic per-prediction evaluation \\
\end{longtable}

Individual predictions in §5 are evaluated on subsets of the 35-model
corpus, and the subset is stated where each is reported: the P2 ordering
recompute uses the \textbf{29 base models} carrying all seven concepts
(§5.2); the P7 scale correlation uses \textbf{26 models} (§5.7); the
chain-depth decay analysis uses \textbf{28 base models} for the
remaining-depth result and \textbf{26 models} for the
unembedding-clustering sub-result (§5.8, the latter's 182 measurements
being 26 × 7); and the P3 width test uses \textbf{22 models} across 6
concepts (§5.8, n = 132). Each subset is the set for which the relevant
artifact exists, not a selection on outcome.

\textbf{The 7-concept corpus is this paper's evaluative paradigm.} The
seven concepts (credibility, certainty, causation, temporal\_order,
sentiment, negation, moral\_valence) are the set on which the
framework's predictions were pre-registered, and every result this paper
computes \emph{itself} --- CAZ detection, the prediction scorecard (§5),
the worked examples (§6), and both figures --- is reported at that C=7
scope. The validation paper {[}Henry, 2026c{]} re-evaluates all seven
predictions on the expanded 17-concept corpus (adding ten later
behavioral and safety concepts), and the GEM handoff results {[}Henry,
2026b{]} use the same 17-concept set; those C=17 figures are
\emph{cited} here as the broader test and, where they differ, supersede
this paper's C=7 assessment (§8).

\begin{center}\rule{0.5\linewidth}{0.5pt}\end{center}

\subsection{2. Relationship to Existing
Work}\label{relationship-to-existing-work}

The CAZ framework operates in the same space as several established
interpretability methods. Each captures different aspects of model
internals; the CAZ contribution is the layer-indexed dynamical view ---
it tracks \emph{when} representations form, rather than only
characterizing \emph{what} they are.

\subsubsection{2.1 Methodological Context}\label{methodological-context}

\textbf{Logit Lens and Tuned Lens.} The most direct methodological
antecedents of the CAZ framework are the logit lens {[}nostalgebraist,
2020{]} and the tuned lens {[}Belrose et al., 2023{]}. Both project
intermediate residual-stream activations into a fixed interpretive space
--- the unembedding matrix for logit lens, a per-layer trained affine
probe for tuned lens --- to produce a layer-by-layer trajectory of the
model's evolving prediction. CAZ shares the core commitment to tracking
representations across depth rather than characterizing a single best
layer, and any claim about layer-indexed concept dynamics necessarily
inherits from this line of work. The distinction is what is being
projected and asked. Logit and tuned lens project onto output-vocabulary
space and ask \emph{what token would this layer predict if the model
stopped here?}; the CAZ framework projects onto concept-contrast
directions obtained from external stimulus pairs, then tracks the
resulting class separation across depth to ask \emph{where in the stack
is this concept being allocated?} Concepts without a canonical
lexicalization --- credibility, refusal, moral valence --- do not
project cleanly onto vocabulary axes and fall outside the natural scope
of logit-lens interpretation, which motivates the contrast-direction
formulation used here. Calibrating the velocity metric against Tuned
Lens in settings where concept and vocabulary are tightly coupled is
left to future work.

\textbf{Sparse Autoencoders (SAEs)} decompose activations at a fixed
layer into interpretable monosemantic features {[}Cunningham et al.,
2023; Bricken et al., 2023{]}. SAEs answer ``what features exist at
layer L?'' The CAZ framework answers ``how does the feature at layer L
relate to the feature at layers L−1 and L+1?'' The approaches are
complementary: SAE features at a given layer are a snapshot; CAZ
tracking reveals which snapshots are part of the same evolving
computation. Engels et al.~{[}2025a{]} decompose SAE ``dark matter'' ---
the unexplained variance in activations --- and find that about half of
the error vector, and more than 90\% of its norm, is \emph{linearly
predictable} from the input activation; the residual that resists linear
prediction is a distinct and much smaller ``nonlinear error'' component.
It is that nonlinear component, not dark matter as a whole, that the CAZ
framework offers a candidate mechanism for: in-progress concept
construction within a CAZ produces transitional representations that are
neither the input feature nor the output feature. These transitional
states may be precisely what resists decomposition at any single layer.

\textbf{Centered Kernel Alignment (CKA)} {[}Kornblith et al., 2019{]}
measures representational similarity between layers or models by
comparing activation kernel matrices. CKA provides a global similarity
score between two representation spaces but does not identify
\emph{which} features are shared or how they evolve across depth. The
CAZ framework tracks individual concept directions layer by layer,
producing a trajectory rather than a scalar comparison. Whether CKA
serves as a complementary validation tool --- high within a CAZ where
the representation is being refined, low at inter-CAZ saddle points
where it is being reallocated --- is left to future work.

\textbf{Linear probing} {[}Belinkov, 2022; Alain \& Bengio, 2017{]}
trains classifiers on activations at each layer to measure concept
presence. Probing accuracy curves are closely related to the separation
metric \(S(l)\) --- both measure how distinguishable two classes are at
a given depth. The CAZ framework adds the velocity metric (rate of
change of separation) and the coherence metric (geometric quality of the
separating direction) --- formalized in §4 --- which together identify
where a concept is present \emph{and} where it is actively being
constructed. Probing also requires training a classifier per layer; the
CAZ metrics are computed directly from activation statistics. Prior work
has also established that this depth-ordered probing view recovers a
syntactic-to-semantic gradient --- shallow layers encode
surface/syntactic features, deeper layers semantic ones {[}Tenney et
al., 2019{]} --- which is the empirical basis for our Prediction 2
(relative cross-architecture ordering, §5.2).

\textbf{Concept activation vectors (TCAV).} Concept directions in
activation space trace to concept-based explanation methods such as TCAV
{[}Kim et al., 2018{]}, which derives a concept vector from a linear
classifier over concept-present vs.~concept-absent activations and
measures a concept's influence on predictions via directional
derivatives. Our concept directions are instead obtained by
difference-of-means over contrastive pairs (cf.~{[}Zou et al., 2023{]};
{[}Marks \& Tegmark, 2024{]}), and our object of study is the
\emph{depth} at which a concept is maximally separable rather than its
sensitivity to a downstream prediction --- so TCAV is an adjacent
concept-based-explanation precursor, not the construction the CAZ
directions derive from.

\textbf{Representation Engineering (RepE)} {[}Zou et al., 2023{]}
extracts concept directions for honesty, morality, power-seeking, and
related concepts via contrastive stimuli, then uses those directions for
monitoring and steering. The CAZ framework bears on RepE's operational
decisions, with the caveat that everything below comes from the
validation paper's geometric ablation protocol and has not been
validated for behavioral steering ({[}Henry, 2026c{]} §6.8). First, the
CAZ profile identifies where a concept is being actively constructed
versus where it is established, offering a candidate basis for choosing
intervention depth. Second, the model's encoding strategy determines
whether layer selection matters at all: in models with redundant
encoding, the concept direction persists across all post-CAZ layers,
suggesting layer choice matters less there; in models with sparse
encoding, the CAZ peak concentrates the measurable signal, and
intervening elsewhere may miss the target. Third, the scored CAZ
detector (§4) reveals that a concept may have multiple candidate layers
--- whether RepE steering applied at a gentle CAZ (§4) produces
different behavioral effects than steering at the dominant peak is an
open question.

\textbf{Activation patching and causal scrubbing.} Activation patching
--- replacing activations at a specified layer and component with those
from a counterfactual run {[}Vig et al., 2020; Meng et al., 2022; Wang
et al., 2023{]} --- is the canonical toolkit for testing whether a
proposed locus genuinely carries a computation. Causal scrubbing {[}Chan
et al., 2022{]} formalizes this as a discipline for rigorously testing
mechanistic hypotheses; path patching {[}Goldowsky-Dill et al., 2023{]}
extends the method to circuit-level attribution. Any claim of the form
``the concept is allocated at layer \(l\)'' is a candidate for
validation under this family of methods, and the ablation protocol used
in the validation paper {[}Henry, 2026c{]} is a direct descendant. The
CAZ contribution is upstream of patching rather than parallel to it: the
velocity-based boundary and score-based region detectors (§4) produce a
principled, concept-specific choice of \emph{which} layers to patch,
replacing exhaustive layer sweeps with a targeted hypothesis about the
allocation zone. The two methods are therefore complementary --- CAZ
identifies candidate layers; activation patching tests whether the
identified layers are causally necessary.

\subsubsection{2.2 Related Empirical
Findings}\label{related-empirical-findings}

\textbf{MLP feed-forward layers as key-value memories.} Geva et
al.~{[}2021, 2022{]} showed that FFN layers function as key-value
memories that incrementally promote concepts into the vocabulary space
across depth. The CAZ metrics (\(S(l)\), \(v(l)\), \(C(l)\)) are defined
on the residual stream and are agnostic to whether promotion happens
through attention or feed-forward paths; connecting CAZ boundaries to
specific FFN write operations is a testable mechanistic question not
pursued here.

\textbf{Manifold interpretability.} Gurnee et al.~{[}2026{]} found
curved manifolds in early layers for character counting and explicitly
called for unsupervised geometric discovery methods. The CAZ framework
provides a formalism for identifying \emph{where} in the layer stack
such manifolds crystallize --- the allocation zone is precisely where
curved manifold structure should be most geometrically coherent.

\textbf{The geometry of refusal.} Arditi et al.~{[}2024{]} and
Wollschläger et al.~{[}2025{]} establish the geometric structure of
refusal --- a single removable direction (abliteration) and
multi-dimensional concept cones, respectively. The CAZ framework extends
this by asking \emph{when} the geometry forms, alongside \emph{what} it
is, and whether that temporal structure bears on the choice of
intervention point (§5.1).

\textbf{Multi-dimensional concept structure.} Engels et al.~{[}2025b{]}
found circular multi-dimensional representations for temporal concepts
that are not decomposable into independent one-dimensional SAE features.
Wollschläger et al.~{[}2025{]} showed refusal occupies polyhedral
concept cones. These findings establish that rich geometric structure
exists; the CAZ framework provides a layer-indexed account of when such
structures crystallize.

\textbf{Concept geometry in production models.} Sofroniew et
al.~{[}2026{]} demonstrate that Claude Sonnet 4.5 develops internal
representations of 171 emotion concepts whose geometry is stable across
model depth (particularly early-middle to late layers) and generalizes
across contexts and behaviors. The representations causally influence
model behavior --- including downstream effects on reward hacking,
sycophancy, and blackmail --- confirmed via activation steering. Concept
vectors are extracted by averaging residual stream activations across
stories per emotion and subtracting the cross-emotion mean; the
resulting representations cluster by semantic similarity with valence
and arousal as primary axes, mirroring human psychological structure.
This result provides direct support for two foundational claims
underlying the CAZ framework: that transformers develop stable,
generalizable geometric representations of human-interpretable concepts
in their residual streams --- representations that hold across diverse
inputs rather than being probe artifacts --- and that this structure is
causally meaningful rather than a mere correlate of surface features
(this paper's own causal evidence for CAZ specifically is discussed in
§7).

\textbf{The Platonic Representation Hypothesis.} Huh et al.~{[}2024{]}
proposed that models trained on different data and architectures
converge on shared representations. The CAZ framework enables a
depth-stratified test of this hypothesis: rather than measuring global
alignment, we can ask whether convergence differs at shallow versus deep
processing stages.

\textbf{Universal neurons.} Gurnee et al.~{[}2024{]} identified
individual neurons in GPT-2 that activate on consistent input features
across five independent training runs --- direct evidence that specific
representational structure is stable under retraining. This is a
neuron-level instantiation of the architectural-convergence phenomenon
the Platonic Representation Hypothesis {[}Huh et al., 2024{]} addresses
at the level of whole representation spaces. The CAZ framework predicts
a depth-stratified extension of both claims: if concepts allocate at
architecturally stable proportional depths, then the layers hosting
universal neurons for a given concept should correspond to that
concept's CAZ peak in proportional terms, and cross-architecture
alignment of concept directions should be strongest when evaluated at
matched CAZ depths rather than at matched absolute layer indices.
Universal-neuron discovery and CAZ-profile matching would then be two
lenses on the same underlying regularity at different granularities.

\begin{center}\rule{0.5\linewidth}{0.5pt}\end{center}

\subsection{3. Background}\label{background}

\subsubsection{3.1 The Residual Stream and Concept
Representation}\label{the-residual-stream-and-concept-representation}

The residual stream formulation {[}Elhage et al., 2021{]} treats each
layer's output as an additive contribution to a shared communication
channel. Attention heads and MLPs read from and write to this stream;
the final residual vector is projected onto the unembedding matrix to
produce logits. This architecture makes layer-by-layer tracking of
concept geometry natural: we can ask, at each layer \emph{l}, how well
the current residual stream separates two contrastive classes.

\subsubsection{3.2 Difference-of-Means and Linear Artificial
Tomography}\label{difference-of-means-and-linear-artificial-tomography}

DoM extracts a concept direction \(V_{\text{concept}} \in \mathbb{R}^d\)
as the normalized difference between class-conditional mean activations
at the chosen layer {[}Zou et al., 2023{]}. Linear Artificial Tomography
(LAT) uses a similar contrastive approach. Both methods produce a single
vector at a single depth --- a precise and useful representation of
where the concept is most geometrically legible. The CAZ framework asks
what additional information about concept formation might be recoverable
from the layers surrounding that peak.

\subsubsection{3.3 Abliteration and Intervention
Depth}\label{abliteration-and-intervention-depth}

Arditi et al.~{[}2024{]} demonstrated that refusal behavior across 13
open-source models is mediated by a single direction removable via
weight orthogonalization (``abliteration''), and report that this
intervention disables refusal with minimal effect on other capabilities.
The CAZ framework motivates a systematic study of intervention depth:
rather than selecting a single layer by hyperparameter search, the CAZ
profile identifies where concepts are being actively constructed versus
where they are established, offering a candidate basis for choosing
intervention points (§5.1 reports the pre-registered mid-CAZ version of
this idea as not supported).

\subsubsection{3.4 The Emerging Geometric
Program}\label{the-emerging-geometric-program}

Gurnee et al.~{[}2026{]}, Engels et al.~{[}2025b{]}, and Wollschläger et
al.~{[}2025{]} --- surveyed in §2.2 --- establish a growing body of
evidence for rich geometric structure in activation space, and their
authors have explicitly called for unsupervised methods to detect and
characterize it. The CAZ framework complements this program by providing
a layer-indexed account of \emph{when} such structures crystallize, in
addition to \emph{where}. The companion GEM paper {[}Henry, 2026b{]}
extends this same question to extraction itself, tracking how a
concept's separating direction rotates within a CAZ before it settles
(§4.4).

\begin{center}\rule{0.5\linewidth}{0.5pt}\end{center}

\subsection{4. The Concept Allocation
Zone}\label{the-concept-allocation-zone-1}

A CAZ is a \textbf{locator}: a coordinate on model depth that brackets
where a concept is assembled --- not the concept itself, and not a
discrete functional module.

Operationally, a CAZ is the segment of the residual stream between two
saddle points of the separation curve \(S(l)\), around a local
separation peak (§4.3). The detector partitions depth into such segments
and scores each; the score grades \textbf{geometric salience}
(Major→Gentle; see Terminology) --- not causal importance (the
validation paper {[}Henry, 2026c{]} §6.4), and not membership: every
segment is a CAZ, and ``strong'' versus ``gentle'' refers to score.
Because the segmentation covers the full depth, a CAZ is
depth-\emph{extended}, not depth-\emph{localized} --- in the limit a
single segment can span the entire network (the GPT-2-XL causation case:
one region across all 48 layers, §6.1) --- so the score says how sharply
the coordinate is drawn, not how much computation the network performs
there.

Where the causal work occurs is a separate question the CAZ does not
answer. It is displaced from the separation peak (the validation paper
{[}Henry, 2026c{]} §6.4), and in redundant-encoding models causal
efficacy extends past the CAZ boundary entirely --- the optimal
intervention point is post-CAZ, not mid-CAZ (§5.1, §5.8). The CAZ marks
the zone; it does not claim to locate the computation within or beyond
it.

The concept is the human label we project onto the geometry. A single
CAZ may host multiple concepts simultaneously --- a depth-indexed analog
of polysemanticity, where a shared geometric resource serves more than
one concept rather than one feature serving more than one meaning
({[}Henry, 2026c{]}) --- and a single concept typically participates in
multiple CAZes across depth (mean 2.20 per concept per model). When we
say ``credibility has a CAZ at layer 10,'' we mean layer 10 falls within
a segment allocated to credibility's geometric expression --- not that
layer 10 ``contains'' credibility.

\clearpage

\subsubsection{Terminology}\label{terminology}

\begin{longtable}[]{@{}
  >{\raggedright\arraybackslash}p{(\columnwidth - 2\tabcolsep) * \real{0.5000}}
  >{\raggedright\arraybackslash}p{(\columnwidth - 2\tabcolsep) * \real{0.5000}}@{}}
\toprule\noalign{}
\begin{minipage}[b]{\linewidth}\raggedright
Term
\end{minipage} & \begin{minipage}[b]{\linewidth}\raggedright
Definition
\end{minipage} \\
\midrule\noalign{}
\endhead
\bottomrule\noalign{}
\endlastfoot
\textbf{CAZ} & A locator --- a coordinate on model depth that brackets
where a concept is assembled. Operationally, a scored saddle-to-saddle
segment of the separation curve (§4.3); not the concept, and not a
functional module \\
\textbf{CAZ Profile} & The full sequence of CAZes for one concept in one
model \\
\textbf{Major CAZ} & CAZ with score \textgreater{} 0.5 --- the highest
composite-salience band; score reflects prominence, coherence, and width
jointly (§4.3), not spatial narrowness --- a Major CAZ can span an
entire network \\
\textbf{Strong CAZ} & CAZ with 0.2 \textless{} score ≤ 0.5 \\
\textbf{Moderate CAZ} & CAZ with 0.05 \textless{} score ≤ 0.2 \\
\textbf{Gentle CAZ} & A low-salience CAZ (score \textless{} 0.05),
invisible to standard peak detection; the validation paper finds these
regions causally active under geometric ablation in most cases (§7) ---
a direct illustration that the score grades salience, not causal
importance \\
\textbf{Embedding CAZ} & CAZ at the earliest stored (transformer-block)
layers, driven by near-lexical/token-level features rather than deep
computation. The pre-transformer embedding output is excluded from
extraction, so this denotes the earliest \emph{block} layers, not the
raw embedding (see §4.1) \\
\textbf{Active CAZ} & CAZ within the transformer layers, driven by
attention and MLP computation. Active CAZes are the primary subject of
this framework \\
\textbf{CAZ score} & Composite metric: (prominence / mean separation)
\(\times\) coherence boost \(\times \sqrt{\text{width}/N}\) (full form
in §4.3) \\
\textbf{Separation \(S(l)\)} & Within-class-scale-normalized centroid
distance between contrastive classes at layer \(l\) \\
\textbf{Coherence \(C(l)\)} & Explained variance ratio of the primary
separating direction at layer \(l\) \\
\textbf{Velocity \(v(l)\)} & Smoothed rate of change of \(S(l)\) across
layers \\
\textbf{Directional Stability \(DS(l)\)} & Layer-to-layer cosine
similarity of the dominant concept direction \(\hat{d}(l)\); near 1.0
means the direction barely moved; a sharp drop marks a handoff event
(§4.2) \\
\end{longtable}

\subsubsection{4.1 Concept Lifecycle}\label{concept-lifecycle}

By tracking the residual stream across model depth, concept formation is
empirically distinguishable as a sequence of allocation regions rather
than a single allocation followed by decay.

\textbf{Early layers (Context and Syntax)}

In early layers, the residual stream primarily resolves local context,
grammar, and surface token relationships. Projecting contrastive
datasets into this space generally produces heavily entangled
activations; the separation metric is near zero --- the contrastive
classes are not yet linearly distinguishable at these depths.

However, the scored detector applied to this paper's own corpus does
find CAZes at the earliest stored layers (0--1) in tens of concept ×
model pairs, concentrated in concepts with strong lexical cues.
Credibility is the clearest case: using an independent method (surface
lexical cues, not the scored detector), the validation paper separately
resolves credibility below 25\% depth in 9 of 28 models ({[}Henry,
2026c{]} §3.3). \textbf{Layer-indexing convention.} Extraction excludes
the model's pre-transformer embedding output (HuggingFace
\texttt{hidden\_states{[}0{]}}): stored layer 0 is the \emph{first
transformer block's} output, stored layer \(l\) is the output of block
\(l{+}1\), and depth percentages are taken over the \(N\) block outputs.
These earliest-layer CAZes therefore sit at the first one or two
transformer blocks, not at the raw embedding. Most are
gentle-to-moderate (score \textless{} 0.1) and plausibly reflect
near-lexical structure already present in early layers ---
concept-associated tokens carrying distinctive early-block geometry, a
property of the tokenizer and training corpus rather than deep
computation; direct token-embedding alignment does not confirm this
account (§5.6). We term these \textbf{embedding CAZes} --- a label for
their near-lexical, earliest-layer character, \emph{not} a claim that
they occur before any transformer processing (the literal embedding
layer is not in the extracted tensor) --- to distinguish them from
\textbf{active CAZes} where the model's attention and MLP computations
allocate geometry to serve a concept. Practitioners using CAZ profiles
should be aware that embedding CAZes may not respond to the same
interventions as active CAZes.

\textbf{CAZ Chain (Concept Allocation)}

As the residual stream deepens, geometric directions are allocated to
concepts in contiguous depth regions --- the CAZes. A persistent
direction in activation space may serve one concept at shallow layers,
be reallocated to a different concept at a mid-depth CAZ, and
reallocated again at a deeper one. For a given concept, the separation
metric \(S(l)\) reflects when that concept holds a strong claim on one
or more geometric directions; a single concept typically participates in
multiple CAZes across depth (mean 2.20 per concept per model under
scored detection), ranging from major CAZes (score \textgreater{} 0.5)
to gentle allocation regions (score \textless{} 0.05; Terminology).

\textbf{Late layers (Logit Projection)}

In the final layers, the model transitions from abstract representation
to concrete next-token prediction. The residual stream is projected
toward the unembedding matrix. Concept geometry may degrade or
re-entangle as abstract directions give way to vocabulary-specific
structure. This is the only phase where separation genuinely decays;
earlier apparent ``decay'' between CAZ peaks may instead reflect
reallocation of the direction to a different concept.

\subsubsection{4.2 Layer-Wise Metrics}\label{layer-wise-metrics}

Let \(h_l^{(i)} \in \mathbb{R}^d\) be the residual stream activation at
layer \(l\) for sample \(i\), and let \(A, B\) be contrastive classes
with conditional means \(\bar{h}_A^{(l)}, \bar{h}_B^{(l)}\) and
within-class covariance matrices \(\Sigma_A^{(l)}, \Sigma_B^{(l)}\).

\textbf{Separation Metric}

We define the separation at layer \(l\) as a
within-class-scale-normalized centroid distance --- a practical
separability proxy in the spirit of Fisher's discriminant analysis
{[}Bishop, 2006, §4.1.4{]}, though, as noted below, not Fisher's
criterion itself:

\[S(l) = \frac{\left\| \bar{h}_A^{(l)} - \bar{h}_B^{(l)} \right\|_2}{\sqrt{\tfrac{1}{2}\left(\operatorname{tr}(\Sigma_A^{(l)}) + \operatorname{tr}(\Sigma_B^{(l)})\right)}}\]

In plain terms: separation asks ``if I gave you a sentence and asked
whether it expresses credibility or not, how easily could you tell from
the model's internal state at this layer?'' A high \(S(l)\) means the
model's activations for credible and non-credible text have moved far
apart relative to how spread out each group is. A low \(S(l)\) means the
two groups are still jumbled together. Tracking \(S(l)\) across layers
reveals where the model begins to ``make up its mind'' about a concept.

Raw centroid distance is misleading when cluster dispersion varies
across layers. Early layers tend toward diffuse, high-variance
representations; normalization by within-class spread corrects for this.
Mahalanobis distance {[}Mahalanobis, 1936{]} would account for full
covariance structure but requires estimating and inverting per-class
covariances, which is ill-conditioned in high-dimensional activation
spaces without shrinkage regularization --- a route we do not pursue
here. Our scalar denominator normalizes only by the \emph{total}
(isotropic) within-class scale,
\(\tfrac{1}{2}(\operatorname{tr}\Sigma_A + \operatorname{tr}\Sigma_B)\),
not by scatter along the discriminant direction. Consequently \(S(l)\)
is \textbf{not} Fisher's discriminant criterion, and is not guaranteed
monotone with linear separability under general (anisotropic)
covariance: two layers with equal \(S(l)\) can differ in Bayes-optimal
separability, and residual streams are known to be anisotropic (a few
rogue dimensions dominate \(\operatorname{tr}\Sigma\)). We adopt it as a
closed-form, training-free separability proxy that is numerically stable
and comparable across layers, not as a Bayes-optimal measure.

\textbf{Concept Coherence}

Separation alone is insufficient: two classes could exhibit identical
centroid separation while one forms a tight cluster and the other a
diffuse cloud. We track Concept Coherence as the explained variance
ratio of the first principal component of the full pooled activation
matrix at each layer:

\[C(l) = \frac{\lambda_1^{(l)}}{\sum_i \lambda_i^{(l)}}\]

where \(\lambda_i^{(l)}\) are the eigenvalues of the covariance of all
activations --- positive and negative class combined --- at layer \(l\).
Concretely, \(C(l)\) is computed by fitting PCA on the concatenated
\([h_A; h_B]\) matrix and reading off the explained variance ratio of
the first principal component. A high \(C(l)\) means the dominant source
of variation across both classes is a single direction --- the
representation is geometrically concentrated. A low \(C(l)\) means
variation is spread across many dimensions --- the representation is
diffuse. A concept is \emph{well-formed} when both \(S(l)\) and \(C(l)\)
are high: the classes are far apart \emph{and} the separating direction
is geometrically clean.

In plain terms: coherence asks ``is the concept encoded as a single
clean direction, or is it smeared across many dimensions?''
\textbf{Precision note}: because \(C(l)\) is fitted on the pooled
(both-class combined) activation matrix, it measures the dominant
direction of total variance, which may correspond to within-class spread
rather than between-class separation. \(C(l)\) is most interpretable in
combination with \(S(l)\): high \(C(l)\) with high \(S(l)\) confirms the
dominant pooled direction is also the class-separating direction; high
\(C(l)\) with low \(S(l)\) indicates activation concentration on a
direction that is not class-discriminative. Figure 2 shows a worked
example of the two curves moving apart: coherence peaks well before the
deepest separation peak and is already declining by the time \(S(l)\)
reaches it.

\textbf{Concept Velocity}

To characterize the rate of geometric divergence between layers, we
compute a smoothed first difference of the separation curve. Because raw
layer-to-layer differences are noisy, we first apply a centered
moving-average of width \(w\) to \(S(l)\) and then take the
adjacent-layer difference:

\[v(l) = \tilde{S}(l) - \tilde{S}(l-1), \qquad \tilde{S}(l) = \frac{1}{w}\sum_{j=-\lfloor w/2 \rfloor}^{\lfloor w/2 \rfloor} S(l+j)\]

where \(\tilde{S}\) is \(S\) box-smoothed over a centered window of
width \(w\). The centered sum as written has
\(2\lfloor w/2 \rfloor + 1\) terms, so it matches the \(1/w\)
normalization only for odd \(w\) --- satisfied by the fixed default
\(w=3\) used throughout this paper (reference implementation:
\texttt{rosetta\_tools.caz.compute\_velocity}). At the endpoints the
window is zero-padded (numpy \texttt{same}-mode convolution), which
mildly attenuates \(\tilde{S}\) within \(\lfloor w/2 \rfloor\) layers of
each boundary --- affecting only the diagnostic velocity trace, not the
CAZ boundaries, which are taken from \(S(l)\) directly. All results in
this paper use the library's fixed default \(w=3\); a depth-adaptive
setting \(w = \max(1, \lfloor N/24 \rfloor)\) is also available but was
not used for the corpus reported here. Velocity is reported as a
trajectory \emph{diagnostic} --- it shows where separation is rising or
falling --- but the CAZ boundaries used throughout this paper are taken
from the peaks and saddle points of \(S(l)\) directly (§4.3,
Multi-Region Detection), not from a velocity threshold.

In plain terms: velocity asks ``is the concept forming right now, or has
it already formed?'' Positive velocity means separation is increasing
--- the model is actively constructing the concept at this layer.
Negative velocity means separation is decreasing --- the concept is
being degraded or reallocated. Zero velocity means nothing is changing;
velocity turns positive as allocation begins and negative as it ends. In
the unimodal single-region case (§4.3) this profile can bound a CAZ
directly; the scored multi-region detector used for the results in this
paper instead places boundaries at the saddle points of \(S(l)\), with
velocity retained as a visualization aid.

\textbf{Directional Stability}

The three metrics above track \emph{how strongly} a concept is
expressed. \(DS(l)\) tracks \emph{which direction} it occupies. Let
\(\hat{d}(l)\) be the normalized dominant direction (dom\_vector) at
layer \(l\) --- the unit vector from negative to positive class
centroid. Directional stability is:

\[DS(l) = \hat{d}(l) \cdot \hat{d}(l-1)\]

\(DS(l)\) near 1.0 means the dominant direction barely moved between
consecutive layers; the model is encoding the concept stably. A sharp
drop marks a \textbf{handoff} --- a layer at which the geometric basis
of the concept shifts substantially. The Geometric Evolution Map (GEM)
framework {[}Henry, 2026b{]} reads the \emph{settled} direction at each
CAZ segment's final layer and evaluates it at the handoff layer --- the
first layer past the segment's saddle boundary; the handoff layer is
derived from that boundary (§4.4), not detected from a drop in
\(DS(l)\). The GEM angular velocity \(\omega(l) = 1 - |DS(l)|\) is the
complement, using the absolute value to treat direction-sign ambiguity
as zero rotation rather than full rotation. The two readouts therefore
differ on a pure sign flip by design --- a sharp drop in signed
\(DS(l)\), but zero rotation in \(\omega(l)\) --- because antipodal
directions span the same one-dimensional axis.

\(DS(l)\) is conceptually independent from \(S(l)\): separation can be
high while the direction is still rotating (during active CAZ assembly),
and separation can be low while the direction is stable (between concept
peaks). The two signals decompose model activity into \emph{what} is
being encoded and \emph{how stably}, respectively.

\subsubsection{4.3 CAZ Boundary Detection}\label{caz-boundary-detection}

\paragraph{Single-Region Detection
(Velocity-Based)}\label{single-region-detection-velocity-based}

When the \(S(l)\) curve is unimodal, CAZ boundaries are derived from the
velocity profile:

\begin{itemize}
\tightlist
\item
  \textbf{CAZ Entry} (\(l_{\text{start}}\)): The first layer where
  \(v(l)\) exceeds a sustained positive threshold
  \(\theta_+ = 0.5 \times \max_l v(l)\), maintained for at least two
  consecutive layers. Note: \(\theta_+\) uses the global maximum of
  \(v(l)\) across all layers of the model, including any
  negative-velocity portions.
\item
  \textbf{CAZ Peak} (\(l_{\text{max}}\)): The layer where \(S(l)\)
  reaches its absolute maximum. This corresponds to the ``best layer''
  of conventional interpretability.
\item
  \textbf{CAZ Exit} (\(l_{\text{end}}\)): The layer where \(v(l)\)
  becomes consistently negative for at least two consecutive layers,
  marking the onset of post-CAZ degradation.
\end{itemize}

The conventional best-layer heuristic extracts \(V_{\text{concept}}\) at
\(l_{\text{max}}\). CAZ-aware extraction uses the full interval
\([l_{\text{start}}, l_{\text{end}}]\).

On the temporal\_order example in Figure 1 below, per-CAZ ablation
confirms both regions are causally active rather than an artifact of the
peak-vs-peak-distal comparison §7 flags as partly circular: suppression
reduces separation by 46.5\% of separation at the shallow CAZ and 77.5\%
at the deep one --- a non-tautological check, since it compares each
region's own ablated state to its own intact state rather than comparing
selected-for-high-separation layers to selected-for-low-separation ones.

\begin{figure}
\centering
\includegraphics{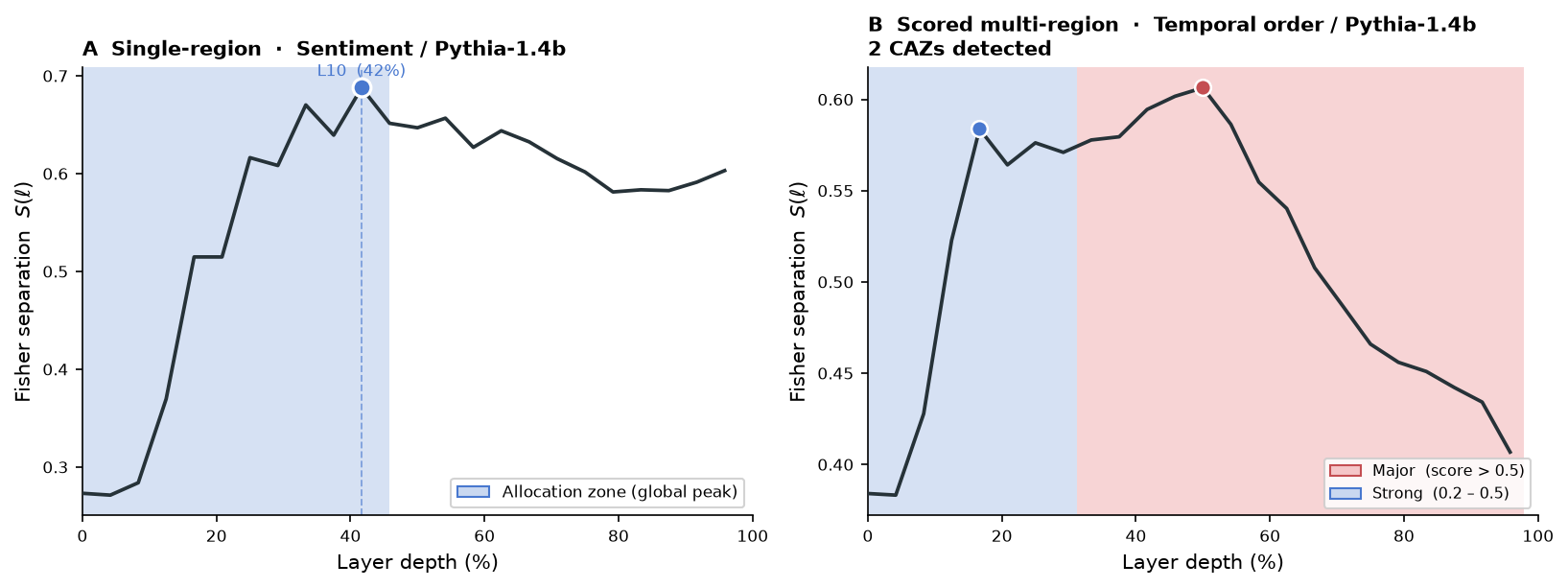}
\caption{Figure 1: Single-region detection (left) versus scored CAZ
profile (right). The single-region detector brackets one allocation zone
around the global separation peak for sentiment in Pythia-1.4B (peak
L10, 42\% depth). The scored detector, applied to temporal\_order in the
same model, resolves two CAZes at geometrically distinct depths (§4.5):
a shallow strong CAZ (L4, 17\% depth, score 0.47) and a deep major CAZ
(L12, 50\% depth, score 0.65) --- the single-region method would bracket
only the deeper peak and miss the shallow one. N=250 contrastive pairs,
\(w=3\) velocity smoothing (diagnostic only), final-token probing,
scored detection defaults (0.5\% prominence floor, 3\% valley-merge
threshold).}
\end{figure}

\paragraph{Multi-Region Detection (CAZ
Profiles)}\label{multi-region-detection-caz-profiles}

Empirical analysis across 35 models reveals that the \(S(l)\) curve is
frequently \textbf{multimodal} --- a single concept can produce multiple
significant local maxima at different depths. In these cases, the
velocity-based boundary detector wraps a single contiguous zone around
the global maximum and is blind to secondary peaks.

The \textbf{CAZ Profile} generalizes the single-region CAZ to a sequence
of allocation regions:

\begin{enumerate}
\def\labelenumi{\arabic{enumi}.}
\tightlist
\item
  Detect all significant local maxima in \(S(l)\) using prominence-based
  peak detection.
\item
  Identify \textbf{saddle points} --- the local minima between
  consecutive peaks --- as natural region boundaries.
\item
  Each region spans from one saddle to the next, with the first region
  starting at layer 0 and the last ending at the final layer.
\end{enumerate}

This segmentation \textbf{tiles the full depth} --- every layer belongs
to exactly one region --- so ``CAZ'' names a segment of the partition,
not a sparse subset of privileged layers; a region's \emph{strength} is
carried by its composite score (below), not by membership. This totality
is a property of \emph{this detector as configured}, not a theorem about
CAZes in general: the detector emits at least the global separation-peak
region as a fallback when no prominence-qualifying peak is found and
applies no score filter, so coverage is complete by construction
(verified across the GEM paper's corpus, 493 concept × model pairs over
29 base models × 17 concepts; {[}Henry, 2026b{]}). The degenerate limit
of this totality is therefore a \emph{single} segment spanning the whole
network (the GPT-2-XL causation case), never zero coverage; a
differently-constructed segmenter need not tile. Note also that the
first and last boundaries are imposed by the model's endpoints rather
than detected from a saddle; in particular the final region always
terminates at the last layer. Because the partition is total, there are
no ``non-CAZ layers'' in an absolute sense: where this paper and its
companions contrast CAZes against a peak-distal baseline (the
peak-suppression enrichment; the gentle-CAZ causal checks), the
comparator is defined \emph{relative to peaks} --- the validation paper
uses layers more than three from any CAZ peak ({[}Henry, 2026c{]} Table
8). The size of that peak-distal comparator varies with model depth, a
dependence the enrichment comparison must account for (see {[}Henry,
2026c{]}).

A CAZ Profile is characterized by: - \textbf{n\_regions}: Number of
distinct allocation regions (1 = unimodal, 2+ = multimodal) -
\textbf{peak region}: The region with the highest peak separation -
Per-region: start, peak, end, width, peak separation, coherence,
rise/fall asymmetry

\paragraph{Scored Detection}\label{scored-detection}

The prominence-based detector requires a threshold to determine which
peaks are significant. A fixed threshold (e.g., 10\% of global max
separation) is arbitrary and risks discarding subtle but causally active
allocation regions. We replace it with a composite scoring system:

\[\text{CAZ score} = \frac{\text{prominence}}{\bar{S}} \times \left(1 + \frac{C(l_{\text{peak}})}{\bar{C}}\right) \times \sqrt{\frac{\text{width}}{N}}\]

where \textbf{prominence} is the scipy peak-prominence value --- the
height of the peak above the higher of its two surrounding saddle points
--- and peaks are required to clear a minimum prominence of 0.5\% of the
global maximum \(S(l)\) (the \texttt{prominence=} selection threshold
passed to \texttt{scipy.find\_peaks}: peaks below it are not detected,
rather than clamped); \(\bar{S}\) is the global mean separation across
all layers; \(C(l_{\text{peak}})\) is the coherence at the peak layer;
\(\bar{C}\) is the global mean coherence across all layers; width is the
region's layer count; and \(N\) is the model's layer count. The three
factors have natural interpretations: the first normalizes prominence by
the baseline signal level; the second (coherence boost) rewards peaks
with above-average geometric organization, with a peak at the model-wide
mean coherence scoring 2.0; the third rewards wider, more sustained
regions while preventing broad shallow regions from dominating. The
0.5\% floor was an initial empirical choice rather than a derived
threshold; post-hoc ablation across detected gentle CAZes finds them
causally active when suppressed --- the validation paper {[}Henry,
2026c{]} confirms this at scale (full breakdown and threshold-dependence
caveat in §7) --- consistent with the floor admitting predominantly
causal regions rather than noise. This is an upper-bound calibration: it
confirms the floor is not too permissive, but does not establish whether
it is too conservative --- regions below the floor may also be causally
active, and lower-bound calibration is open work. The scored detector is
the primary detection method; the fixed-threshold detector is retained
for backward compatibility.

\textbf{Detection parameters.} Three further parameters govern region
detection and are held at their defaults throughout this paper: a
minimum peak separation of 2 layers (\texttt{min\_peak\_distance}); a
valley-merge threshold (\texttt{min\_valley\_depth\_frac\ =\ 0.03}) that
fuses adjacent regions whose separating valley is shallower than 3\% of
the global peak --- on a merged region, prominence is recomputed as the
peak height above the higher merged endpoint rather than the raw scipy
saddle value, so the formula above applies to unmerged regions; and an
optional architecture-calibrated re-ranking
(\texttt{functional\_caz\_score}, which reweights the geometric score by
attention-paradigm ablation priors) that exists in the library but is
\textbf{not} applied to this paper's reported region counts. A
reimplementer should set these to the stated defaults to reproduce the
results here.

\subsubsection{4.4 Multi-Layer Concept
Extraction}\label{multi-layer-concept-extraction}

The CAZ framework motivates the question: given that a concept assembles
across multiple layers, where is the best layer to extract its probe
direction? Three candidate methods are:

\begin{enumerate}
\def\labelenumi{\arabic{enumi}.}
\item
  \textbf{Delta PCA}: PCA on layer-to-layer residual deltas
  \(\Delta h_l = h_l - h_{l-1}\) within
  \([l_{\text{start}}, l_{\text{end}}]\). This captures what each layer
  \emph{adds} --- the construction process itself.
\item
  \textbf{Windowed PCA}: PCA on raw activations \(h_l\) across
  \([l_{\text{start}}, l_{\text{end}}]\). This captures the cumulative
  concept direction as it evolves through the allocation zone.
\item
  \textbf{Single-layer (baseline)}: The standard dom\_vector at
  \(l_{\text{max}}\).
\end{enumerate}

GEM continues this line of reasoning: the companion paper {[}Henry,
2026b{]}, published simultaneously with this work, tracks the
directional trajectory of a concept through the residual stream and
shows that concept probe directions undergo substantial rotation
\emph{within} the CAZ, not settling into a stable encoding until a
characteristic depth past the CAZ boundary --- the \emph{handoff layer},
derived from the same saddle-based segmentation defined in §4.3. Probing
at the handoff layer is measurably more precise than peak-layer
extraction in 341/493 concept × model trials (69.2\%; model-level
Wilcoxon W=356, N=29, one-sided). Excluding the two Gemma-2 models
(whose single-direction ablation readout is known to fail on their
distributed encoding rather than indicating a negative result, §7)
raises this to 326/459 = 71.0\%. A depth-matched control finds the
settling contribution beyond depth-of-extraction is weak (263/493 =
53.3\% of pairs, not significant under any reported per-model
aggregator): this is a depth effect, not a settling effect --- probing
deeper helps, independent of whether the deeper layer is where the
direction specifically settles. The advantage holds regardless of
attention architecture (MHA vs.~GQA); an early architecture-dependent
reading did not survive the expanded corpus (``Changes from Version
1''). The GEM paper has run the matched-layer delta-PCA comparison
(near-equivalent to the difference-of-means probe, no layer effect
established); windowed PCA remains distinct and unrun. Full results and
the cohort breakdown are reported in {[}Henry, 2026b{]}.

\subsubsection{4.5 Sub-Representations}\label{sub-representations}

When a concept's \(S(l)\) curve is multimodal, the dom\_vector (the
difference-in-means direction of §4.2 --- the unit vector between class
centroids, not a principal component) at each peak defines a distinct
linear direction. Empirical measurement across 34 multimodal concept ×
model pairs (6 concepts --- sentiment collapses to zero multimodal pairs
at N=250; full breakdown in {[}Henry, 2026c{]} Table S6, §J) shows these
directions are \textbf{geometrically distinct}: per-concept mean cosine
between the shallow and deep peak dom\_vectors falls in the range
0.12--0.41 --- well above the random baseline (\textbar cos\textbar{}
\(\approx\) 0.02 for random unit vectors in these dimensions) but well
below identity. The per-concept counts are small (under the N=250-sample
scored detector, as few as 4 models for credibility), so the range
endpoints should be interpreted cautiously. The two peaks are not the
same direction at different amplitudes --- they are candidate distinct
linear features that both separate the same contrastive classes.

This implies that a single human concept label (``credibility'',
``negation'') maps to \textbf{multiple sub-representations} at different
processing depths. Interpretive evidence suggests:

\begin{itemize}
\tightlist
\item
  \textbf{Shallow sub-representations} form near the embedding layers,
  plausibly related to lexical cues (concept-associated words) ---
  though a direct token-embedding alignment test does not confirm this
  (§5.6).
\item
  \textbf{Deep sub-representations} form in mid-to-late layers, likely
  driven by compositional processing (contextual inference, scope,
  pragmatic reasoning).
\end{itemize}

Whether these are genuinely distinct sub-representations or a single
direction rotating gradually across depth is not resolved by the present
data. Layer × layer cosine similarity matrices show block-diagonal
structure, and adjacent-layer cosine can dip at the saddle point (to as
low as 0.35 in some models) --- suggestive of a regime boundary --- but
an independent boundary check in the companion validation paper finds no
reliable co-location between separation saddle points and
direction-rotation events ({[}Henry, 2026c{]}). We therefore describe
the two peaks as candidate distinct encodings without claiming an abrupt
phase transition.

If sub-representations within a single model are geometrically distinct,
the natural next question is whether they correspond \emph{across}
models: does a shallow sub-representation in one architecture align with
a shallow sub-representation in another, and do deep sub-representations
likewise pair? A monolithic single-rotation alignment between
architectures would predict no such depth structure; a depth-stratified
picture predicts that matched-depth pairs align more strongly than
mismatched-depth pairs. This is the testable claim formalized as
Prediction 5 (§5.5); a proper cross-architecture test requires alignment
machinery beyond this paper's scope and is deferred to dedicated
follow-up work.

\begin{center}\rule{0.5\linewidth}{0.5pt}\end{center}

\subsection{5. Testable Predictions}\label{testable-predictions}

The CAZ framework generates seven predictions that are in principle
falsifiable with existing open-weight models and standard
interpretability tooling. Predictions were formulated from the
theoretical framework prior to running any analysis; the broad
phenomenon of multimodal allocation was observed early --- preliminary
profiles showed that some models produce multiple peaks --- and this
informed the framework's inclusion of scored detection; but the full
quantitative scope (multimodal being the norm rather than the exception,
mean 2.20 CAZes per concept per model) was not established until the
validation pipeline ran. P4 in particular was formulated under the
assumption that single-peak models were the primary case and multimodal
profiles were exceptional. The specific directional predictions below
were committed before per-concept results were examined.

Predictions P1--P7 were pre-specified on 2026-04-05 and published at
https://waypoint.henrynet.ca/research/concept-assembly-zone/CAZ\_Framework.pdf
prior to running the full validation pipeline; they were not submitted
to a formal external registry such as OSF. Verdicts were recorded after
those analyses ran.

Verdicts use three categories: \textbf{Supported} (prediction holds),
\textbf{Partially supported} (direction correct, stated mechanism or
magnitude wrong), and \textbf{Not supported} / \textbf{Not testable as
stated} (specific claim failed or premise invalidated). Predictions that
cannot be given a clean verdict at this paper's scope are instead marked
\emph{Exploratory}, \emph{Indeterminate}, or \emph{Deferred} (defined in
§8). Findings that emerged from investigating prediction failures ---
including results that held but were not pre-specified --- are reported
separately in §5.8 to keep the pre-specified record clean.

\subsubsection{5.1 Optimal Ablation Depth}\label{optimal-ablation-depth}

\textbf{Prediction 1}: The suppression-to-damage ratio varies
systematically with intervention depth relative to the CAZ.

For a given concept, extract the concept direction at each layer via
DoM. Apply orthogonal projection at each layer \(l\) independently.
Measure: (a) behavioral suppression rate on targeted prompts, (b) KL
divergence from the unmodified model on unrelated prompts. Plot the
ratio (a)/(b) as a function of layer.

\textbf{Status: Not supported.} The prediction specified mid-CAZ as the
optimal intervention point. Empirical results {[}Henry, 2026c{]} falsify
this: in models with redundant encoding (Pythia, GPT-2, OPT), the
concept direction persists throughout the post-allocation residual
stream, making suppression layer-invariant; capability damage is lowest
at late layers, so the optimal point is \emph{post-CAZ}, not mid-CAZ.
The specific claim failed. At its full 17-concept corpus scope the
validation paper's own scorecard rates this prediction \emph{partially
supported} --- the region-level claim holds while peak selection fails
(§8). What investigating the failure revealed --- a two-regime encoding
structure that the prediction did not anticipate --- is reported in
§5.8. (Ablation coverage: §7.)

\subsubsection{5.2 Cross-Architecture CAZ
Ordering}\label{cross-architecture-caz-ordering}

\textbf{Prediction 2}: CAZ boundaries are concept-specific but show
consistent relative ordering across architectures.

Different concepts should have different CAZ windows within the same
model. However, the \emph{relative} ordering of those windows --- as a
fraction of total model depth --- should be consistent across
architectures. Absolute depth percentages may be family-specific, but
relative concept ordering should be consistent.

\textbf{Status: Not supported.} A from-scratch recompute of this paper's
own 7-concept ordering claim, against the 29 base models in the
extraction corpus that carry all seven concepts, does not reproduce the
originally published tendency. The recompute uses the current
scored-detection methodology (0.5\% prominence floor; depth read at each
concept's \emph{dominant CAZ region}, ranked by composite score per the
finalized CAZ definition) --- the same scored detector the validation
paper uses, but not the same dominance convention: the validation
paper's §3 ordering reads depth at the tallest-separation-peak region,
while composite-score dominance is the convention of its §6.4. Median
per-model \(\tau = +0.05\) against the grand-mean consensus ordering (17
of 29 models positive, 12 negative; Wilcoxon signed-rank, two-sided
\(p = 0.54\) --- not significant), with per-model \(\tau\) spanning
\(-0.71\) to \(+0.62\), roughly symmetric around zero rather than
uniformly positive. This originally published figure (median
\(\tau=0.473\), 26/26 models positive, \(p=1.49\times10^{-8}\)) does not
reproduce under the current pipeline; the exact methodology (detector
variant, prominence floor, self-inclusion handling) that produced the
original number is not fully recoverable from the published text alone,
and the 7-concept ordering is in any case superseded by the validation
paper's own from-scratch 17-concept analysis. That analysis ({[}Henry,
2026c{]} §3.1--3.2, 28 base models) finds a real but moderate tendency
(median \(\tau=0.404\), 27 of 28 positive with gpt2-medium exactly zero
and none negative, Wilcoxon \(p=1.49\times10^{-8}\)) --- read as
\emph{partially supported} there. The dominance convention accounts for
part of the gap from that reading: on this paper's own C = 7 / 29-model
data, reading depth at the tallest-separation-peak region --- the
convention the validation paper's §3 ordering uses --- rather than at
the composite-score dominant region raises the median to
\(\tau = +0.43\) (supported range). We report the composite-score result
(\(\tau = +0.05\), n.s.) as primary because composite-score dominance is
the finalized CAZ convention, and disclose the metric sensitivity here
rather than adopt the more favorable measure. At this paper's own
7-concept / 29-model scope the ordering prediction is \textbf{not
supported}: it does not reproduce, and the point estimate is near zero
(median \(\tau \approx +0.05\), not significant). The validation paper's
broader-corpus reading reaches \emph{partially supported} (§8), but this
paper's own data do not support the prediction. The consensus ordering
(shallow \(\to\) deep) spans a syntactic-to-behavioral gradient across
17 concepts from specificity to exfiltration (mean depths
21.4\%--81.0\%; {[}Henry, 2026c{]} §3). Detailed per-model breakdown in
{[}Henry, 2026c{]}.

\subsubsection{5.3 CAZ Width and Concept
Abstraction}\label{caz-width-and-concept-abstraction}

\textbf{Prediction 3}: CAZ width correlates with concept abstraction
level.

More abstract concepts (e.g., ``trustworthiness,'' ``moral valence'')
should have wider CAZ windows than concrete ones (e.g., ``negation,''
``plurality''), because abstract concepts require more iterative
construction across attention layers. This is testable by comparing
\(l_{\text{end}} - l_{\text{start}}\) for concepts at different levels
of semantic abstraction as operationalized by, for example, depth in
WordNet or scores on standard concreteness rating datasets.

\textbf{Status: Exploratory --- see §5.8.} With \(n = 6\) concepts and a
researcher-assigned abstraction ranking, the sample is too small to
treat this as a tested prediction; the result is reported in §5.8 as a
hypothesis-generating finding. The validation paper's full 17-concept
run reaches \emph{partially supported} (§5.8, §8).

\subsubsection{5.4 Post-CAZ Degradation as Logit
Interference}\label{post-caz-degradation-as-logit-interference}

\textbf{Prediction 4}: Post-CAZ re-entanglement correlates with
unembedding matrix structure.

The degradation of clean concept geometry in late layers is not noise
but a structural consequence of preparing the residual stream for logit
projection. Concepts whose associated vocabulary tokens are
distributionally similar in the unembedding space --- close in embedding
distance --- should show more post-CAZ degradation than concepts with
distributionally distinct vocabulary. This would explain why some
concepts retain clean geometry into late layers (their vocabulary is
well-separated) while others degrade early (their vocabulary clusters).

\textbf{Status: Not testable as stated.} P4 was formulated before the
full quantitative scope of multimodal allocation was established --- at
that point, multi-peak profiles were seen in some models but not yet
known to be the norm. The prediction assumes a single post-CAZ decay
region, which is a unimodal premise. Multimodal allocation (mean 2.20
CAZes per concept per model; {[}Henry, 2026c{]}) invalidates that
premise: under multiple allocation regions there is no single post-CAZ
region against which to evaluate the unembedding correlation. (Whether
the inter-peak decline reflects reallocation rather than degradation is
a separate and unresolved question, §4.1; the verdict does not depend on
it.) The prediction is not falsified by the data --- its precondition is
not met. A post-hoc analysis prompted by this failure is reported in
§5.8.

\subsubsection{5.5 Depth-Stratified Representational
Convergence}\label{depth-stratified-representational-convergence}

\textbf{Prediction 5}: Cross-architecture alignment is depth-matched.

When a concept has multiple allocation regions, the sub-representation
at a given processing depth should align more strongly with the
corresponding-depth sub-representation in other architectures than with
a different-depth sub-representation. Specifically, after Procrustes
rotation, cosine(shallow\_A, shallow\_B) \textgreater{}
cosine(shallow\_A, deep\_B) and cosine(deep\_A, deep\_B) \textgreater{}
cosine(deep\_A, shallow\_B).

\textbf{Status: Deferred --- outside this paper's scope.} The prediction
concerns cross-architecture Platonic convergence: whether CAZ
sub-representations align by processing depth across architectures.
Evaluating it properly requires a dedicated cross-architecture alignment
methodology with appropriate null controls. That machinery, and the
prior question of whether any depth-matched signal is concept-specific
rather than merely layer-specific, is the subject of dedicated follow-up
work and is not adjudicated here: this framework paper neither confirms
nor refutes P5. (v1 scored this prediction \emph{Supported} on a
peak-to-peak reading; this paper does not have the machinery to
re-adjudicate that reading, so it withdraws the verdict and defers the
question rather than restating it.)

\textbf{Dataset note.} The validation-paper-sourced C=17 figures this
paper cites elsewhere --- the P2 ordering τ = 0.404 in §5.2 and the P3
width correlation ρ = 0.307 in §5.8 --- are computed on the multi-model
consensus Rosetta Concept Pairs dataset (14 generators --- six model
families across four AI labs; {[}Henry, 2026c{]}). This paper's
\emph{own} 7-concept computations (the τ null recompute in §5.2, the
figures, and the worked examples in §6) use the single-model Claude
Sonnet 4.6 concept dataset described in §7.

\subsubsection{5.6 Lexical vs.~Compositional
Sub-Representations}\label{lexical-vs.-compositional-sub-representations}

\textbf{Prediction 6}: Shallow peaks encode lexical features; deep peaks
encode compositional features.

The dom\_vector at the shallow allocation peak should correlate with
token embedding vectors for concept-associated words (e.g.,
``reliable'', ``dubious'' for credibility). The dom\_vector at the deep
peak should show lower correlation with token embeddings and higher
dependence on multi-token contextual patterns.

\textbf{Status: Not supported.} Token embedding probing --- cosine
similarity between peak dom\_vectors and concept-relevant token
embeddings --- yields near-zero values (\(\sim 0.02\)) at \emph{both}
the shallow and deep peaks. A paired two-sided Wilcoxon signed-rank test
across the multimodal concept × model pairs (§4.5; \(n = 34\)) finds no
difference between them (\(p = 0.82\)): the effect the prediction
requires --- shallow peaks more token-aligned than deep --- is absent,
and both peaks are near-orthogonal to raw token embeddings regardless.
This is a small exploratory test at the 7-concept scale. The
lexical/compositional distinction may operate at a higher level of
abstraction than direct embedding alignment --- the shallow feature
could depend on token identity through multi-layer composition rather
than literally pointing toward any single token's embedding vector.
Testing this requires alternative experimental designs: per-token
position attribution, attention knockout, or probing classifiers trained
at each peak.

\subsubsection{5.7 Multi-Modality as Architectural
Property}\label{multi-modality-as-architectural-property}

\textbf{Prediction 7}: Multi-modality prevalence is determined by
architecture, not scale.

The fraction of concepts showing multimodal \(S(l)\) curves should vary
more between architectural families (attention mechanism, activation
function, training data) than between scales within a family.

\textbf{Status: Indeterminate.} The scale correlation is near zero
(\(\rho = 0.11\), \(p = 0.63\), \(n = 26\) models; the validation paper
notes this correlation has not been recomputed on the full 28 {[}Henry,
2026c{]} §5.7), but at this \(n\) the test is underpowered --- the 95\%
CI for \(\rho\) includes values up to \textasciitilde0.47, so absence of
a significant correlation is not evidence for the prediction. The
qualitative family-level differences are real (Qwen 2.5: deep, prominent
bimodality, valley depths 26--36\%; Gemma 2: subtle structure below the
10\% prominence threshold), but these are descriptive observations, not
a formal test. Resolving P7 requires either a within-family scale ladder
spanning a wider parameter range than the current Pythia ladder (8
sizes, 70M--12B --- enough breadth in principle, but underpowered here
for a formal architecture-vs-scale contrast) or a formally
pre-registered architecture-level test against a stated alternative;
both are open work {[}Henry, 2026c{]}.

\subsubsection{5.8 Findings Beyond the
Predictions}\label{findings-beyond-the-predictions}

Three results either emerged from investigating prediction failures or
were not pre-specified to begin with. They are reported here separately
to distinguish hypothesis-driven from data-driven findings.

\textbf{Two-regime encoding structure (from investigating P1).}

P1 predicted mid-CAZ ablation would be optimal; it was not (§5.1).
Investigating why revealed a confound the prediction did not anticipate:
encoding strategy determines whether layer choice matters at all. In
models with redundant encoding (Pythia, GPT-2, OPT), the concept
direction persists throughout the post-allocation residual stream, so
suppression is layer-invariant. In models with sparse encoding (Qwen,
Gemma), layer specificity is greater and CAZ location matters more. The
two regimes are confounded with architecture vintage in this corpus
(redundant-encoding models are older/absolute-position; sparse-encoding
models are newer/RoPE-based), so encoding strategy specifically, as
opposed to architecture family generally, is not isolated as the
operative variable. This is a productive falsification: the two-regime
model is more explanatorily complete than the original prediction.
Whether the CAZ profile \emph{predicts} which regime applies before
ablation --- as opposed to describing it post hoc --- is not established
here: no held-out predictor was tested, and this is offered as a
hypothesis for future validation {[}Henry, 2026c{]}.

\textbf{Chain-depth decay signal (post-hoc analysis prompted by P4).}

P4's precondition --- a single post-CAZ decay region --- does not hold
under multimodal allocation. After establishing this, we examined a
redefined target: degradation from the \emph{final} CAZ peak to the last
layer. The validation paper's full-corpus test finds that remaining
depth predicts decay (\(r = -0.265\), \(p = 4.08\times10^{-9}\); 476
measurements, 28 base models × 17 concepts; mean decay ratio 0.848) and,
on the subset of 7 original concepts with hand-curated token lists (182
measurements, 26 models), unembedding token clustering shows a weak
effect (\(r = 0.173\), \(p = 0.019\)) {[}Henry, 2026c{]}. Neither set of
measurements is independent --- each shares models across concepts ---
so the pooled \(p\)-values are anticonservative; the effect sizes and
their direction, not the \(p\)-values, are the evidentiary content.
These results are consistent with the intuition behind P4 and are
reported here as exploratory findings; they do not rescue P4 as stated.

\textbf{Width-abstraction correlation (P3, exploratory).}

\textbf{Operationalization note}: §5.3 proposed operationalizing
``abstraction level'' via depth in WordNet or scores on standard
concreteness rating datasets. The executed test uses a
researcher-assigned abstraction ranking instead. The proposed external
validation (WordNet depth or concreteness ratings) was not carried out;
the ranking was established by the author before examining per-concept
widths.

Excluding credibility (bimodal, high variance), CAZ width correlates
with researcher-assigned abstraction rank, with affective and epistemic
concepts showing wider CAZes than relational and syntactic ones. This
paper's original 6-concept test (\(r = 0.294\), \(n = 132\)
concept-model pairs = 6 concepts × 22 models, permutation \(p = 0.003\)
on concept-level means) --- nominally significant, but scored
\emph{exploratory} in the scorecard because \(n = 6\) concepts is too
small to constitute a test (§5.3, §8) --- has since been superseded by
the validation paper's full 17-concept run across 45 models (28 base + 9
instruct-tuned + 8 additional scale/frontier models --- a wider corpus
than the standard 28-base-model ablation set), which reaches the
\emph{partially supported} level: Spearman \(\rho = 0.307\),
\(p = 3.7 \times 10^{-18}\) over \(n = 765\) CAZ sweeps {[}Henry,
2026c{]}. Two caveats are carried from that validation-paper result: the
abstraction ranking is author-assigned (external-ranking replication
remains open), and the pooled \(p\) treats sweeps that share models and
concepts as independent, so the modest effect size (\(\rho = 0.307\))
and its direction --- not the pooled \(p\) --- are the evidentiary
content. The full 17-concept activation corpus (45 models) is available
at the Rosetta Activations dataset {[}Henry, 2026, dataset{]}.

\begin{center}\rule{0.5\linewidth}{0.5pt}\end{center}

\subsection{6. Proof of Concept}\label{proof-of-concept}

To demonstrate that the CAZ metrics and detection methods produce
meaningful results, we present a minimal example on GPT-2-XL (48 layers,
1.5B parameters) using 7 concepts with 250 contrastive pairs each.

\subsubsection{6.1 CAZ Detection}\label{caz-detection}

The separation curve \(S(l)\) for credibility in GPT-2-XL peaks at layer
45 (94\% depth) with \(S = 1.17\) --- the strongest signal among the 7
concepts tested (N=250 contrastive pairs per concept, \(w=3\) velocity
smoothing, final-token probing). The scored detector identifies 2 CAZes
for this concept under default detection settings (0.5\% prominence
floor, 3\% valley-merge threshold; §4.3): a moderate early CAZ at L9
(19\%, score=0.17) and the dominant peak at L45 (94\%, major,
score=0.84) --- the dominant mode is the \emph{deep} pole of
credibility's bimodal profile (§4.5 multimodality), not a mid-network
assembly event.

Across all 7 concepts in this single model, dominant allocation peaks
span 27--94\% depth (negation L13, temporal\_order L21, certainty and
moral\_valence L23, sentiment L24, causation L25, credibility L45).
Negation has two detected strong CAZes at L13 (27\%, score=0.44) and L19
(40\%, score=0.41) --- close enough in score that which one is
``dominant'' is a near-tie, reflecting the shallow-to-deep
sub-representation transition described in §4.5. Causation shows a
single detected region at L25 under the current scored detector (a
previously-observed shallow secondary region near L13 does not clear the
detection threshold in the current corpus). The tendency for syntactic
and relational concepts to peak earlier than epistemic ones holds for
most of the set, though credibility is a clear exception in this model
--- consistent with P2's ordering tendency (§5.2) not replicating
cleanly at this paper's original 7-concept scope, though the
broader-corpus picture is more favorable (§5.2).

\subsubsection{6.2 Scored Detection Reveals Hidden
Structure}\label{scored-detection-reveals-hidden-structure}

Lowering the detection threshold from 10\% to 0.5\% (scored detection)
increases the number of detected CAZes from 7 to 10 in this single model
--- credibility, negation, and temporal\_order each reveal secondary
allocation regions invisible under the 10\% threshold. The additional
regions are not noise: the validation paper {[}Henry, 2026c{]} finds
them causally active at scale (§7). The additional CAZes here correspond
to interpretable sub-representation transitions (§4.5). GPT-2-XL is not
the richest example of multimodal structure; models with more pronounced
allocation dynamics (e.g., Pythia-160M, which shows a mean of 3.1 CAZes
per concept across the 7 evaluated concepts) provide more vivid
demonstrations --- and the full cross-model pattern is reported in
{[}Henry, 2026c{]}. Whether regions below the 0.5\% floor are also
causally active remains untested.

\begin{figure}
\centering
\includegraphics{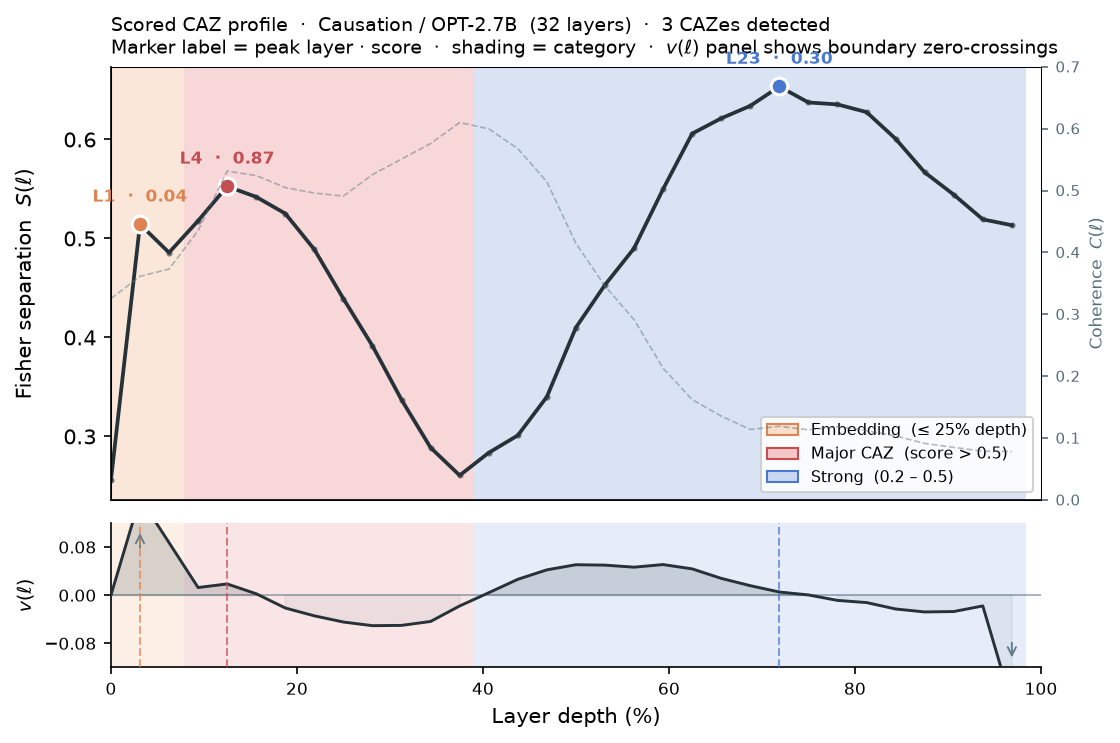}
\caption{Figure 2: Scored CAZ profile for causation in OPT-2.7B (32
layers). Three CAZes detected: a gentle embedding CAZ at layer 1 (3\%
depth, score=0.04), a major CAZ at layer 4 (12\% depth, score=0.87), and
a strong CAZ at layer 23 (72\% depth, score=0.30). The dashed line
(right axis) is Coherence \(C(l)\): it peaks near the saddle around 25\%
depth and is already declining by the deep separation peak at L23 ---
separation and coherence move independently (§4.2), and this profile is
a direct illustration. N=250 contrastive pairs, \(w=3\) velocity
smoothing (diagnostic only; boundaries are from the scored
separation-curve detector), final-token probing, scored detection
defaults. Depth percentages use the paper's \(l/N\) convention (§4.1),
matching the rendered axis ticks.}
\end{figure}

\subsubsection{6.3 Scope of Validation}\label{scope-of-validation}

The framework has been evaluated across 35 models from 8 architectural
families (Pythia, GPT-2, OPT, Qwen 2.5, Gemma 2, Llama 3.x, Mistral,
Phi) spanning 70M to 14B parameters. Full empirical results ---
including multi-family scale ladders, structural analysis, and dark
matter quantification --- are reported in the validation paper {[}Henry,
2026c{]}. The reference implementation is provided as rosetta\_tools
v1.3.1 {[}Henry, 2026, software{]}.

\begin{center}\rule{0.5\linewidth}{0.5pt}\end{center}

\subsection{7. Limitations}\label{limitations}

Known limitations include:

\textbf{Synthetic contrastive data bias}

This is the most significant limitation of the current work. All 7
concept datasets used for validation were generated by a single model
(Claude Sonnet 4.6). Every CAZ location, every separation score, and
every depth estimate in this paper is measured against Claude's
definition of each concept. If Claude's concept boundaries differ
systematically from a human consensus definition, then what we are
mapping is Claude's ontology projected onto other models, not a
universal property of the models themselves. Multi-model consensus pair
generation is addressed in the validation paper: Henry {[}2026c{]} §C
replicates the peak-depth analysis on four architecturally diverse
models using both the original single-generator pairs and the
\textasciitilde1,400-pair RCP consensus pool (14 generators --- six
model families across the Anthropic, Google, OpenAI, and Mistral labs),
finding that the category-level ordering is preserved under both, while
per-concept peak depths shift substantially in places (median
\textbar Δ\textbar{} ≈ 9.4 pp). The framework methodology is independent
of the data source; only the specific empirical results are at risk.

\textbf{Pair fidelity}

Human validation of the RCP pairs (the validation paper, Henry
{[}2026c{]} §2.2/§8.8; a 31-rater crowd study, 540 analysis-grade
ratings) confirms the corpus is high-fidelity for the large majority of
concepts --- 14 of 17 validate at ≥76\%, though at the pair level only
66\% of rated pairs are clean on this first pass --- but not uniformly
so. A minority, moral\_valence most severely (no clean pair of ten),
were built as opposite-pole (antonym) contrasts rather than
presence-versus-absence contrasts, so their negatives still express the
concept. Because CAZ extraction recovers whatever geometry the pairs
make the model separate on, a bad pair still solidifies a stable,
causally-active direction --- just the wrong axis (for moral\_valence,
moral \emph{polarity} rather than moral-valence \emph{presence}).
Per-concept results for these low-fidelity concepts should be read as
geometry over the pole contrast, not concept presence; the aggregate
results are carried by the high-fidelity majority and are unaffected.
This is a property of the pair data, not the method --- automated
separability cannot catch it (a probe separates antonym pairs
perfectly), and only human fidelity validation does. Henry {[}2026c{]}
§8.8 develops the point.

\textbf{Model scale}

Full-corpus validation at N=250 extended coverage to the larger models
--- Pythia-12B, Gemma-2-9B, and Llama-3.1-8B --- that were deferred from
earlier reduced-scale runs. Frontier-scale models (70B+) were not
evaluated in this paper's own CAZ runs; whether CAZ \emph{structure} and
the depth-ordering predictions hold at that scale is untested here.

\textbf{Smoothing sensitivity}

Under the scored detector used for all results here, CAZ boundaries are
set by the separation curve's peak-prominence floor and valley-merge
threshold (defaults 0.5\% and 3\%), not by the velocity smoothing width
\(w\) or a velocity threshold: because the detector segments \(S(l)\)
directly, recomputing velocity at any smoothing window leaves the region
counts and boundaries unchanged. The prominence and valley-merge
defaults produce consistent results across 35 models but have not been
formally validated against ground-truth concept boundaries.

\textbf{Linearity assumption}

The separation metric assumes the concept manifold is approximately
linearly separable. For concepts with curved or multi-dimensional
structure {[}Gurnee et al., 2026; Engels et al., 2025b{]}, kernel-based
or topological metrics may be required. This limitation is shared with
most of the current interpretability literature.

\textbf{Token position dependence}

The framework probes a single fixed token position (the final token by
default) and treats the resulting layer-wise separation curve as
representative of the concept. Zhao et al.~{[}2025{]} demonstrate this
directly for safety-relevant concepts: harmfulness encodes at the last
instruction token while refusal encodes at the last post-instruction
token --- distinct positions for distinct concepts within the same
prompt. CAZ boundaries derived from a single fixed token position may
not generalize to concepts that encode elsewhere. No systematic
sensitivity analysis on alternate token positions was conducted in this
work; this is a known gap rather than an oversight, and position-varied
replication is a concrete direction for follow-up.

\textbf{Scope: semantic concepts only}

The CAZ framework applies to concepts that are assembled through
transformer layers via the allocation dynamics described here.
Tokenization-level concepts --- representations tied to byte-level
patterns, script detection, BPE boundary artifacts, or sub-word
orthography --- do not undergo CAZ assembly in the same sense: they are
largely established at the embedding layer and do not exhibit the
velocity peaks, saddle-point transitions, or cross-architecture
alignment that define the CAZ phenomenon. All seven concepts in the
current validation corpus (credibility, certainty, causation,
temporal\_order, sentiment, negation, moral\_valence) are semantic, and
the framework's cross-model observations in this paper are scoped to
that class. Extending CAZ analysis to tokenization-level representations
requires a separate treatment of how such features behave under the
layer-indexed metrics defined in §4.2.

\textbf{Causal validation coverage}

Ablation testing in this paper covers 16 of 35 models; the validation
paper {[}Henry, 2026c{]} extends ablation coverage to all 28 base models
(on the seven original concepts; the gentle-CAZ ablation table does not
cover the ten later concepts) and confirms that 98\% of gentle CAZes
clear the validation paper's 20-percentage-point operating threshold,
versus 28\% of peak-distal layers --- though the enrichment over
peak-distal regions is threshold-dependent, ranging from ≈1.7× at a
5-percentage-point threshold to ≈20× at 50 ({[}Henry, 2026c{]} §8.7/§H),
and the comparison is not fully independent of the selection criterion
that defines CAZ vs.~peak-distal layers in the first place (circularity
caveat below). These causal claims are geometric (ablation-induced
separation reduction); a single-layer behavioral pilot confirms the
ablated directions are functionally active on next-token predictions
(direction-specific in 27/28 models) but finds peak-vs-matched-control
suppression indistinguishable from chance ({[}Henry, 2026c{]} §6.8) ---
so the framework's \emph{peak-placement} causal support is geometric,
not yet behavioral. The 6 instruct-tuned variants in this paper's corpus
are not covered by {[}Henry, 2026c{]}'s base-model ablation sweep.

\textbf{SAE residual cross-validation}

§1 notes that CAZ regions may correspond to the nonlinear component of
SAE reconstruction error --- the residual that resists \emph{linear}
prediction from the input activation at a fixed layer, a minority of the
error norm {[}Engels et al., 2025a{]}. The mechanism is plausible:
in-progress concept construction within a CAZ produces transitional
representations that are neither the input feature nor the output
feature. This hypothesis has not been tested in this work. Direct
cross-validation --- measuring whether gentle CAZ layers predict
elevated SAE unexplained residual in the same model --- is a concrete
open direction.

\textbf{Construct and predictive limitations}

Several results in this paper and its companions cut against the
framework; they are consolidated here rather than left to the reader to
assemble. (1) \emph{Prediction scorecard}: of the seven pre-registered
predictions, none is confirmed (full breakdown, §8). (2) \emph{Score
does not track causal importance}: the CAZ score predicts geometric
salience but not ablation impact --- the validation paper documents
higher-scored shallow CAZes with negligible suppression alongside
lower-scored deep CAZes producing \textasciitilde30\% reduction
({[}Henry, 2026c{]} §6.1/§6.6), so the Major/Strong/Moderate/Gentle
bands are a salience ordering, not a causal ranking. (3) \emph{Gemma-2
readout failure}: Gemma-2 produces strong-range CAZ scores while
single-direction ablation at the Fisher peak registers only weakly (mean
0.367, against 0.503 at CAZ peaks and 0.140 at peak-distal layers
corpus-wide), and its cross-architecture ordering is not reliably
estimated ({[}Henry, 2026c{]} §6.9). The validation paper shows this is
what single-direction intervention \emph{must} produce on a distributed
code regardless of the underlying causal structure, so those cells are
uninformative about Gemma-2 rather than evidence of causal inertness ---
its CAZes are unmeasured by this instrument, not shown non-causal
(activation patching, which needs no direction estimate, recovers the
concept near-completely there). (4) \emph{Methodological
reproducibility}: the v1 ordering statistic (§5.2) is not fully
recoverable from the published method description alone. (5)
\emph{Null-model coverage}: the gentle-CAZ permutation null was
evaluated on 5 of 28 base models, with the remainder assumed to
generalize ({[}Henry, 2026c{]}). (6) \emph{Circularity}: the construct
is defined by the separation metric \(S(l)\) (Fisher-style but not
Fisher's criterion, §4.2), whose numerator is the class-centroid
difference, and validated by projecting out that same direction and
measuring the resulting drop in centroid separation, which annihilates
the between-class mean at the ablated layer by construction. The
non-tautological content is only what propagates to the \emph{final}
layer. Relatedly, peak-vs-peak-distal comparisons contrast layers
selected for \emph{high} separation against layers selected for
\emph{low} separation, so some positive gap is expected from the
selection criterion alone; these causal claims should be read against
that baseline, and the one non-circular test available (the behavioral
peak-vs-matched-control pilot) is null ({[}Henry, 2026c{]} §6.8).

\textbf{Pre-registration and instrument design}

The seven predictions were pre-specified by the author on 2026-04-05 and
published at \texttt{waypoint.henrynet.ca} prior to running the
validation, but they were \emph{not} deposited with an external registry
(e.g., OSF), so the artifact carries no independent timestamp. More
substantively, the measurement instrument is not fully independent of
the test it feeds: the broad phenomenon of multimodal allocation was
observed early in this corpus and informed the framework's adoption of
scored detection (§5), so the detector that supplies the predictions'
inputs was in part designed on the same data the predictions are
evaluated against. The validation paper's from-scratch re-evaluation on
the expanded corpus {[}Henry, 2026c{]} is the appropriate check against
this (§8).

\begin{center}\rule{0.5\linewidth}{0.5pt}\end{center}

\subsection{8. Conclusion}\label{conclusion}

The Concept Allocation Zone framework provides a methodology for
tracking how concepts form across transformer depth --- complementing
the ``best layer'' view with a dynamical view of concept allocation.

The key contributions are:

\begin{enumerate}
\def\labelenumi{\arabic{enumi}.}
\tightlist
\item
  \textbf{Three layer-wise metrics} (Separation, Coherence, Velocity)
  that characterize concept formation as a process, not a point.
\item
  \textbf{Scored detection} that reveals a spectrum of allocation
  regions from major to gentle, replacing binary thresholds with
  continuous scoring.
\item
  \textbf{The CAZ-is-not-a-concept distinction} --- CAZes are
  depth-extended allocation segments (residual-stream regions bounded by
  separation-curve saddles) in which the model organizes geometry to
  serve concepts. Multiple concepts share CAZes; single concepts
  participate in multiple CAZes across depth.
\item
  \textbf{Sub-representation tracking} across depth: within-model
  dom\_vectors at shallow and deep peaks are not the same direction
  (cosine 0.12--0.41 with block-diagonal cosine similarity structure
  across layers, §4.5) --- whether this reflects genuinely distinct
  sub-representations or one direction rotating gradually across depth
  is not resolved by the present data (§4.5). Whether these
  sub-representations correspond \emph{across} architectures --- the
  cross-architecture depth-matching question (P5) --- is deferred to
  dedicated follow-up work (§5.5).
\end{enumerate}

\textbf{Verdict definitions.} \emph{Supported}: prediction holds.
\emph{Partially supported}: direction correct, stated mechanism or
magnitude wrong. \emph{Not supported}: specific claim failed. \emph{Not
testable as stated}: prediction's precondition invalidated by the data.
\emph{Exploratory}: result exists but sample too small to constitute a
test. \emph{Indeterminate}: the test as run is underpowered and does not
resolve the prediction either way. \emph{Deferred}: the prediction is
outside this paper's evaluative scope and is assessed in dedicated
follow-up work. Results arising from investigating prediction failures
are in §5.8, not this table. The seven predictions are evaluated
individually; no family-wise multiple-comparisons correction is applied
at this paper's level (the validation paper applies one, {[}Henry,
2026c{]}).

\begin{longtable}[]{@{}
  >{\raggedright\arraybackslash}p{(\columnwidth - 4\tabcolsep) * \real{0.1304}}
  >{\raggedright\arraybackslash}p{(\columnwidth - 4\tabcolsep) * \real{0.4783}}
  >{\raggedright\arraybackslash}p{(\columnwidth - 4\tabcolsep) * \real{0.3913}}@{}}
\toprule\noalign{}
\begin{minipage}[b]{\linewidth}\raggedright
\#
\end{minipage} & \begin{minipage}[b]{\linewidth}\raggedright
Prediction
\end{minipage} & \begin{minipage}[b]{\linewidth}\raggedright
Verdict
\end{minipage} \\
\midrule\noalign{}
\endhead
\bottomrule\noalign{}
\endlastfoot
P1 & Mid-CAZ ablation yields optimal suppression-to-damage ratio & Not
supported¶ \\
P2 & CAZ boundaries show consistent relative ordering cross-architecture
& Not supported \\
P3 & CAZ width correlates with abstraction level & Exploratory† \\
P4 & Post-CAZ degradation correlates with unembedding matrix structure &
Not testable as stated‡ \\
P5 & Cross-architecture alignment is depth-matched & Deferred§ \\
P6 & Shallow peaks encode lexical features; deep peaks encode
compositional & Not supported \\
P7 & Multi-modality prevalence is architectural, not scale-dependent &
Indeterminate \\
\end{longtable}

¶See §5.8 (two-regime encoding structure). †See §5.8 (width-abstraction
correlation). ‡See §5.8 (chain-depth decay post-hoc analysis).
§Cross-architecture depth-matched convergence is outside this framework
paper's scope and is evaluated in dedicated follow-up work; it is not
scored here (§5.5). v1's \emph{Supported} verdict rested on a
peak-to-peak reading this paper does not re-adjudicate; the verdict is
withdrawn and the question deferred.

\textbf{Scope of these verdicts.} These are the framework paper's
per-prediction assessments at its original 7-concept evaluation scope.
The validation paper {[}Henry, 2026c{]} re-evaluates all seven
predictions on its full 17-concept, 28-base-model corpus (§5); where the
two differ on scope --- P1, P2, and P3 there reach \emph{partially
supported} --- the validation paper's full-corpus evaluation supersedes
the preliminary verdict here; on the remaining scored predictions the
two papers agree (P4 not testable as stated, P6 not supported, P7
indeterminate). P5 is deferred rather than scored: the
cross-architecture convergence question is outside this paper's scope
and is the subject of dedicated follow-up work.

None of the seven pre-registered predictions is confirmed, and we
present that as an honest result rather than a failure of the framework.
The predictions were committed before the validation corpus existed; a
framework whose purpose is to make concept formation measurable, and
which states falsifiable predictions and reports that they do not
survive contact with the data, has done its job. The framework's
empirical footing is first-party and independent of the scorecard: the
scored detector and its three metrics, the prevalence of multimodal
allocation, the gentle-CAZ category, and --- established in the
validation paper {[}Henry, 2026c{]} --- that the detected allocation
structures are causally real in the geometric sense its protocol
measures, read against the circularity, threshold-dependence, and
behavioral-null caveats set out in §7. The most consequential revision
the data required is the replacement of the single-peak assumption with
multi-peak allocation: an \emph{observed} finding that fed back into the
detector design (§5), not a pre-registered prediction --- it falsified
prediction P4's single-peak premise and broadened the framework's
descriptive scope. The same pattern runs through the scorecard:
investigating prediction P1's failure surfaced the two-regime encoding
structure (§5.8), and the scored detector's lower threshold surfaced the
gentle-CAZ category --- the falsifications are where the framework's
richest current content came from. Full empirical results are reported
in {[}Henry, 2026c{]}.

The reference implementation is available as rosetta\_tools v1.3.1
{[}Henry, 2026, software{]} (DOI 10.5281/zenodo.20361433), an
open-source Python library providing the full CAZ extraction, alignment,
ablation, and feature tracking pipeline described in this paper; the
analysis, figure, and reproduction scripts are Rosetta\_Analysis v1.1.0
(DOI 10.5281/zenodo.21583139), which imports that library, and the
umbrella index for the series is at
https://github.com/jamesrahenry/Rosetta. Pre-extracted model activations
for all corpus models across 17 concepts are available as the Rosetta
Activations dataset at
https://huggingface.co/datasets/james-ra-henry/Rosetta-Activations (DOI
10.57967/hf/9725).

\begin{center}\rule{0.5\linewidth}{0.5pt}\end{center}

\subsection{Changes from Version 1}\label{changes-from-version-1}

This version corrects and retracts claims made in v1 (arXiv:2605.24856).
A reader holding v1 should treat the following as superseded. Changes
are grouped by \emph{why} the value moved, because that distinction
matters: some numbers changed because the corpus was rebuilt or grew,
and others because v1 was wrong.

\textbf{Verdict changes.} Two of the seven pre-registered predictions
changed verdict, and the change is consequential: v1 reported one
prediction supported outright and one partially supported; \textbf{v2
confirms none} --- one is deferred, the other is not supported at this
paper's scope.

\begin{itemize}
\tightlist
\item
  \textbf{P5 (cross-architecture alignment is depth-matched): Supported
  → deferred.} v1 read 563/563 positive trials as confirming the
  pre-specified peak-to-peak claim. Properly evaluating
  cross-architecture Platonic convergence --- including whether any
  alignment is anchored to CAZ \emph{peaks}, as that reading assumed ---
  needs dedicated cross-architecture alignment machinery and null
  controls that are outside this framework paper's scope. Rather than
  restate v1's verdict, P5 is \textbf{deferred to dedicated follow-up
  work} and is neither confirmed nor refuted here (§5.5).
\item
  \textbf{P2 (consistent cross-architecture CAZ ordering): Partially
  supported → Not supported.} v1 reported median τ = 0.473 (26/26 base
  models positive, p = 1.49 × 10⁻⁸). A from-scratch recompute under the
  current scored-detection pipeline, over the 29 base models carrying
  all seven concepts, does not reproduce it: median τ = +0.05 (17/29
  positive, p = 0.54). The exact v1 methodology is not fully recoverable
  from the published description. The validation paper's independent
  17-concept analysis reaches \emph{partially supported} and supersedes
  at that scope, but at this paper's own C = 7 scope the prediction is
  not supported (§5.2).
\end{itemize}

\textbf{Claims withdrawn.} Each was asserted in v1 and is no longer
supported:

\begin{itemize}
\tightlist
\item
  \textbf{The instruction-tuning / alignment-training finding.} v1 (§1)
  asserted that instruction tuning ``does not uniformly shift concept
  allocation to shallower depths; the direction and magnitude of change
  depend on the base model's existing concept geometry,'' and offered it
  as a CAZ-based metric of alignment-training distortion. The empirical
  claim is withdrawn to future work: the validation paper profiles
  base/instruct pairs (Supplementary §B) but no systematic base→instruct
  analysis was carried out. Only the in-principle metric remains,
  explicitly deferred (§1).
\item
  \textbf{``Direct evidence of bandwidth reuse'' (§4.1).} v1 called the
  four layer positions each hosting six of seventeen concept
  eigenvectors ``direct evidence of bandwidth reuse.'' Downgraded: the
  validation paper flags this as a suggestive post-hoc observation
  lacking a permutation null, not an established convergence (§4.1).
\item
  \textbf{The MHA/GQA handoff-cohort finding (§4.4).} v1 reported a
  pronounced, statistically significant architecture split on the
  23-model GEM corpus --- MHA models preferred handoff extraction at
  85\% versus 29\% for GQA (Fisher's exact \(p=0.022\)) --- and stated
  ``the CAZ profile predicts which case applies before extraction.'' On
  the expanded 29-model corpus this does not hold: MHA and GQA prefer
  handoff at statistically indistinguishable rates (76\% vs.~67\%,
  \(p=0.462\)). The handoff-precision advantage itself is not withdrawn
  (it strengthens, 259/391 → 341/493 trials), but the
  architecture-dependence claim is: it was a small-corpus artifact, not
  a property of attention mechanism (§4.4).
\item
  \textbf{Causation's bimodal split (§6.1).} v1's GPT-2-XL worked
  example reported causation splitting across a shallow region near L13
  and its dominant peak at L25. Under the current scored detector and
  the true-N=250 corpus, causation shows a single detected region at
  L25; the shallow secondary region no longer clears the detection
  threshold. The list of concepts with secondary regions in §6.2 changes
  accordingly (causation → temporal\_order).
\item
  \textbf{The sub-representation ``phase transition'' (§4.5).} v1
  described the shallow-to-deep transition at a CAZ's saddle point as
  ``abrupt, not gradual\ldots{} indicating a phase transition between
  distinct encoding regimes.'' An independent boundary check (validation
  paper) finds no reliable co-location between separation saddle points
  and direction-rotation events --- indistinguishable from a permuted
  baseline. v2 withdraws the phase-transition characterization and
  describes the two peaks only as candidate distinct encodings.
\item
  \textbf{``The CAZ profile predicts which regime applies'' (§5.8).} The
  two-regime encoding structure (redundant- vs.~sparse-encoding models,
  discovered investigating P1's failure) is a real, retained finding,
  but v1's implicit claim that the CAZ profile could predict which
  regime applies \emph{before} ablation is run is not established: no
  held-out predictor was tested, and the two regimes are confounded with
  architecture vintage in this corpus.
\end{itemize}

\textbf{Corrections --- v1 was wrong or under-specified, independent of
corpus size.}

\begin{longtable}[]{@{}
  >{\raggedright\arraybackslash}p{(\columnwidth - 6\tabcolsep) * \real{0.2500}}
  >{\raggedright\arraybackslash}p{(\columnwidth - 6\tabcolsep) * \real{0.2500}}
  >{\raggedright\arraybackslash}p{(\columnwidth - 6\tabcolsep) * \real{0.2500}}
  >{\raggedright\arraybackslash}p{(\columnwidth - 6\tabcolsep) * \real{0.2500}}@{}}
\toprule\noalign{}
\begin{minipage}[b]{\linewidth}\raggedright
quantity
\end{minipage} & \begin{minipage}[b]{\linewidth}\raggedright
v1
\end{minipage} & \begin{minipage}[b]{\linewidth}\raggedright
v2
\end{minipage} & \begin{minipage}[b]{\linewidth}\raggedright
why
\end{minipage} \\
\midrule\noalign{}
\endhead
\bottomrule\noalign{}
\endlastfoot
Mean CAZes per concept per model & 3.4 & \textbf{2.20} & v1 used the
pre-N=250-rebuild corpus (mislabeled ≈ 200); recomputed on the true N =
250 corpus (§4.1) \\
Credibility worked example (§6.1) & peak \textbf{layer 29 (60\%), S =
1.05}; dominant region score 0.37 & \textbf{layer 45 (94\%), S = 1.17};
dominant region score 0.84 & true-N=250 rebuild; the dominant mode is
now the deep pole of the bimodal profile \\
Negation worked example (§6.1) & peaks at L17 (35\%, score 0.38) and L13
(28\%, score 0.44) & peaks at L19 (40\%, score 0.41) and L13 (27\%,
score 0.44) --- near-tie for dominance & true-N=250 rebuild \\
Gentle-CAZ causal impact & ``\textbf{93--100\%} of cases'' &
\textbf{98\%} at the 20-pp threshold (enrichment ≈1.7× at 5 pp → ≈20× at
50) & v1's band was the 16-model subset; recomputed on the validation
paper's full 28-base-model sweep (§4.3) \\
Concept depth range (P2) & 21.8\%--84.7\% & \textbf{21.4\%--81.0\%} &
validation-paper recompute at C = 17 \\
Velocity smoothing (captions, §6) & k = ⌊N/24⌋ (captions) and k = 2 (§6)
& fixed default \textbf{w = 3} & the reported corpus uses the library
default; CAZ boundaries come from the scored detector, not velocity
(§4.2) \\
§4.5 sub-representation sample & 46 concept×model pairs, 7 concepts,
cosine range 0.156--0.433 & 34 pairs, 6 concepts (sentiment collapses to
zero multimodal pairs at N=250), cosine range 0.12--0.41 & true-N=250
rebuild \\
Abliteration capability effect (§3.3) & ``independent
replications\ldots{} KL divergence 3.16--5.71\ldots{} the intervention
also affects general model capabilities'' & Arditi et al.~``report that
this intervention disables refusal with minimal effect on other
capabilities'' & v1 mischaracterized the source's own reported result;
corrected to what Arditi et al.~actually report \\
SAE ``dark matter'' (§1, §2.1) & \textasciitilde50\% of the SAE
reconstruction error treated as unexplained ``dark matter'' CAZ may
correspond to & Engels et al.~find \textgreater90\% of that error is
\emph{linearly predictable}; only the small residual \emph{nonlinear}
component is the candidate CAZ correspondence & v1 overstated the size
of the phenomenon being explained \\
``Embedding CAZ'' (Terminology, §4.1) & implied the literal embedding
layer & denotes the first transformer block's \emph{output}; the
pre-transformer embedding is excluded from extraction & precision fix
--- affects how every Embedding CAZ reference should be read \\
Gurnee et al.~{[}2026{]} citation (§2.2) & described the curved-manifold
finding as \textbf{mid}-layer & corrected to \textbf{early}-layer,
matching the source & citation-accuracy correction \\
Consensus-pair ordering check (§7) & ``concept ordering is preserved
across all four test architectures'' & category-level ordering
preserved, but \textbf{per-concept peak depths shift substantially in
places} (median \textbar Δ\textbar{} ≈ 9.4 pp) & validation paper's
fuller RCP analysis added a caveat v1 didn't have \\
\end{longtable}

\textbf{Figures --- both changed subject model and were regenerated.}

\begin{itemize}
\tightlist
\item
  \textbf{Figure 1 (right panel):} v1 showed \emph{credibility in
  Qwen2.5-0.5B}; v2 shows \emph{temporal\_order in Pythia-1.4B}. ``Black
  hole'' terminology → ``Major.''
\item
  \textbf{Figure 2:} v1 showed \emph{negation in OPT-2.7B} (two
  ``black-hole'' CAZes plus a velocity panel); v2 shows \emph{causation
  in OPT-2.7B} (three scored CAZes, no velocity panel).
\end{itemize}

\textbf{Changes attributable to the corrected roster.} The reported
model count is \textbf{34 → 35} (the corpus is 29 base + 6 instruct;
v1's ``8 instruct'' overcounted --- only four families ship instruct
variants). Validation-paper ablation coverage is \textbf{26 → 28} base
models. The GEM handoff corpus is \textbf{23 → 29} base models (raising
the headline handoff-precision trial count from 259/391 to 341/493). The
P7 Pythia scale ladder is \textbf{7 → 8} sizes (70M--12B). The Rosetta
Activations dataset now covers \textbf{33 → 46} language models (40 with
full per-model analysis, 6 frontier models CAZ-only). The v1 hardware
naming (L4/H200 GPUs, §7) was removed; GPU-hours were not logged
(Disclosures).

\textbf{New disclosures since v1 --- not corrections, but not in v1
either.} A reader relying on v1's causal picture should know v2 adds:
(1) a \textbf{pair-fidelity audit} (§7) finding several concepts,
moral\_valence most severely, were built as antonym/opposite-pole
contrasts rather than presence-vs-absence, so their geometry reflects
polarity rather than the intended concept; (2) a \textbf{behavioral-null
result} (§7): a single-layer pilot finds peak-vs-matched-control
suppression indistinguishable from chance, so the framework's causal
support is established only in the geometric sense, not yet
behaviorally; (3) a consolidated \textbf{``Construct and predictive
limitations''} subsection (§7) --- CAZ score doesn't track causal
importance, the Gemma-2 single-direction readout failure, a circularity
concern in the \(S(l)\) construct, and the gentle-CAZ null model having
been checked on only 5 of 28 models; (4) a \textbf{``Pre-registration
and instrument design''} self-critique (§7): the scored detector was in
part designed on the same data the predictions are evaluated against.
None of these change a v1 number --- they're additions v1 did not have
and could not have anticipated (most were added by the validation paper,
{[}Henry, 2026c{]}) --- but they materially change how strongly the
causal claims in this paper should be read.

Full edit-level provenance is in \texttt{V2\_CHANGELIST.md} in the
accompanying repository.

\begin{center}\rule{0.5\linewidth}{0.5pt}\end{center}

\subsection{Appendix: Common
Definitions}\label{appendix-common-definitions}

\emph{This appendix collects the canonical cross-paper definitions
shared by the Rosetta Program's Concept Allocation Zone (CAZ) framework
(this paper), Geometric Evolution Maps (GEM) {[}Henry, 2026b{]}, and CAZ
Validation {[}Henry, 2026c{]}. It is a compact reference for a reader
working from a single paper in isolation --- each concept's full
derivation, motivation, and supporting argument live in the paper cited
alongside it, not here.}

\textbf{CAZ --- Concept Allocation Zone.} (Developed fully in §4 of this
paper.)

\begin{quote}
A CAZ is a locator: a coordinate on model depth that brackets where a
concept is assembled --- not the concept itself, and not a discrete
functional module. Operationally, a CAZ is the segment of the residual
stream between two saddle points of the separation curve \(S(l)\),
around a local separation peak. The detector partitions depth into such
segments and scores each; the score grades geometric salience
(Major→Gentle) --- not causal importance, and not membership.
\end{quote}

Three governing rules: (1) \emph{No membership threshold} --- every
segment is a CAZ; ``strong'' vs.~``gentle'' is score, never a binary
cut. (2) \emph{Depth-extended, not depth-localized} --- the segmentation
tiles the full depth by construction; in the degenerate limit a single
segment can span the entire network. (3) \emph{A CAZ does not locate
causation} --- score is geometric salience, not causal importance; the
causal work can be displaced from the separation peak and can sit
outside the segment entirely.

\textbf{GEM --- Geometric Evolution Map.} (Developed fully in the
companion paper {[}Henry, 2026b{]} §3.)

\begin{quote}
A GEM is the trajectory record of one concept's difference-of-means
direction across one CAZ segment. The settled direction is its terminal
reading, not a co-equal component.
\end{quote}

Components: \textbf{window} --- the layer span of the CAZ segment;
\textbf{trajectory} --- the per-layer dominant directions across that
span, each estimated independently; \textbf{settled direction} --- the
trajectory's reading at the segment's final layer; \textbf{handoff
layer} (\(L_H\)) --- \(\min(\text{end}+1, N-1)\), the first layer
outside the segment, \emph{derived} from the segment boundary and never
independently detected by a rotation criterion.

\textbf{Atlas.} (Developed fully in {[}Henry, 2026b{]} §3.)

\begin{quote}
A concept's atlas is the set of its GEMs in a model --- one GEM per CAZ
segment. An ordinary collection of maps: no transition structure between
them is implied. An atlas is a container, not an operational object ---
it carries no defined rule for combining its GEMs into a single
concept-level answer; any concept-level result applies its own
aggregation rule, which must be stated wherever such a result is
reported.
\end{quote}

\textbf{How they fit together.} CAZ locates; GEM charts; the handoff is
where a settled direction is first evaluated on activations it was not
estimated on, and it sits outside the segment that produced it.

\textbf{Vocabulary --- use these, not the alternatives}

\begin{longtable}[]{@{}
  >{\raggedright\arraybackslash}p{(\columnwidth - 4\tabcolsep) * \real{0.3333}}
  >{\raggedright\arraybackslash}p{(\columnwidth - 4\tabcolsep) * \real{0.3333}}
  >{\raggedright\arraybackslash}p{(\columnwidth - 4\tabcolsep) * \real{0.3333}}@{}}
\toprule\noalign{}
\begin{minipage}[b]{\linewidth}\raggedright
term
\end{minipage} & \begin{minipage}[b]{\linewidth}\raggedright
means
\end{minipage} & \begin{minipage}[b]{\linewidth}\raggedright
do not say
\end{minipage} \\
\midrule\noalign{}
\endhead
\bottomrule\noalign{}
\endlastfoot
CAZ / segment & a scored saddle-to-saddle segment; the locator & ``the
zone where the concept lives''; ``depth-localized'' \\
GEM & one segment's trajectory record & ``the GEM'' meaning a whole
concept; ``assembly event'' \\
atlas & the set of a concept's GEMs in a model & ``map'' (collides with
the M in GEM); ``manifold'' \\
settled direction & the trajectory's terminal reading, at segment's end
& ``the GEM probe''; ``the direction at \(L_H\)'' \\
handoff layer & segment.end + 1, derived & ``detected where rotation
ceases''; ``the handoff within the zone'' \\
caz\_score & geometric salience, graded & ``significance''; a membership
threshold \\
peak-distal & the comparator layer/region away from a CAZ peak &
``non-CAZ'' --- the tiling is total, so there is no non-CAZ layer in an
absolute sense \\
\end{longtable}

\emph{Full provenance and the working argument behind these definitions:
\texttt{DEFINITIONS.md} and
\texttt{papers/shared/CAZ\_DEFINITION\_OF\_RECORD.md} /
\texttt{GEM\_DEFINITION.md} in the accompanying repository.}

\begin{center}\rule{0.5\linewidth}{0.5pt}\end{center}

\subsection{Acknowledgments}\label{acknowledgments}

The author acknowledges the support of TELUS, specifically the Chief AI
Office, the AI Accelerator, and the Chief Security Office.

Thanks to Ivey Chiu, Steve Pearson, and Krista Hickey for helpful
discussions and guidance.

The author acknowledges computational and academic support from the
Vector Institute for Artificial Intelligence.

Thanks to Ilya Grishchenko and David Lie (University of Toronto) for
helpful discussions.

Claude (Anthropic) contributed substantially to this work. Specifically:
iterative discussion of validation methodology design (null experiment
selection, the split-calibration construction-artifact test); manuscript
drafting and structural editing; and code review for the rosetta\_tools
extraction pipeline. The framework's predictions (§5) and all final
interpretive judgements were made by the author.

\textbf{Self-evaluation.} All predictions, experiments, analysis, and
verdicts in this paper are the work of a single author; independent
replication would carry more evidential weight (data and code are
released to enable it). The double-exposure risk the validation paper
flags applies here --- several predictions share one source with their
operationalization, and the scored detector was tuned in part on the
evaluation corpus (§7, ``Pre-registration and instrument design'') ---
so the affirmed results should be read against §7's circularity caveat:
the metric's non-tautological content is only what propagates to the
final layer, and the one non-circular behavioral test available returned
a null result.

\textbf{Compute and licenses.} GPU-hours and energy for extraction were
not systematically logged. The corpus models carry a mix of licenses ---
several (Llama-3.x, Gemma-2) are released under non-OSI community
licenses with use restrictions --- and users should consult each model's
license; the released artifacts (\texttt{rosetta\_tools},
\texttt{Rosetta\_Analysis}, and the activation dataset) are under their
stated open licenses (see §8).

\textbf{Competing interests.} The author is an employee of TELUS
Communications Inc.; this research was conducted independently of that
role, and the author declares no competing interests arising from it.
TELUS and the Vector Institute are acknowledged for their support (see
Acknowledgments); neither had any role in the study's design, data
collection, analysis, interpretation, or the decision to publish.

\begin{center}\rule{0.5\linewidth}{0.5pt}\end{center}

\subsection{References}\label{references}

\begin{itemize}
\item
  Alain, G., \& Bengio, Y. (2017). Understanding intermediate layers
  using linear classifier probes. \emph{arXiv preprint
  arXiv:1610.01644}. https://arxiv.org/abs/1610.01644
\item
  Arditi, A., Obeso, O., Syed, A., Paleka, D., Panickssery, N., Gurnee,
  W., \& Nanda, N. (2024). Refusal in language models is mediated by a
  single direction. \emph{arXiv preprint arXiv:2406.11717}.
  https://arxiv.org/abs/2406.11717
\item
  Belinkov, Y. (2022). Probing classifiers: Promises, shortcomings, and
  advances. \emph{Computational Linguistics}, 48(1), 207--219.
\item
  Belrose, N., Ostrovsky, I., McKinney, L., Furman, Z., Smith, L.,
  Halawi, D., Biderman, S., \& Steinhardt, J. (2023). Eliciting latent
  predictions from transformers with the tuned lens. \emph{arXiv
  preprint arXiv:2303.08112}. https://arxiv.org/abs/2303.08112
\item
  Bishop, C. M. (2006). \emph{Pattern Recognition and Machine Learning}.
  Springer.
\item
  Bricken, T., Templeton, A., Batson, J., Chen, B., Jermyn, A., Conerly,
  T., Turner, N., Anil, C., Denison, C., Askell, A., Lasenby, R., Wu,
  Y., Kravec, S., Schiefer, N., Maxwell, T., Joseph, N., Tamkin, A.,
  Nguyen, K., McLean, B., Burke, J. E., Hume, T., Carter, S., Henighan,
  T., \& Olah, C. (2023). Towards monosemanticity: Decomposing language
  models with dictionary learning. \emph{Transformer Circuits Thread},
  Anthropic.
  https://transformer-circuits.pub/2023/monosemantic-features/index.html
\item
  Chan, L., Garriga-Alonso, A., Goldowsky-Dill, N., Greenblatt, R.,
  Nitishinskaya, J., Radhakrishnan, A., Shlegeris, B., \& Thomas, N.
  (2022). Causal scrubbing: A method for rigorously testing
  interpretability hypotheses. \emph{AI Alignment Forum}, December 2022.
  https://www.alignmentforum.org/posts/JvZhhzycHu2Yd57RN/causal-scrubbing-a-method-for-rigorously-testing
\item
  Cunningham, H., Ewart, A., Riggs, L., Huben, R., \& Sharkey, L.
  (2023). Sparse autoencoders find highly interpretable features in
  language models. \emph{arXiv preprint arXiv:2309.08600}.
  https://arxiv.org/abs/2309.08600
\item
  Elhage, N., Nanda, N., Olsson, C., Henighan, T., Joseph, N., Mann, B.,
  Askell, A., Bai, Y., Chen, A., Conerly, T., DasSarma, N., Drain, D.,
  Ganguli, D., Hatfield-Dodds, Z., Hernandez, D., Jones, A., Kernion,
  J., Lovitt, L., Ndousse, K., Amodei, D., Brown, T., Clark, J., Kaplan,
  J., McCandlish, S., \& Olah, C. (2021). A mathematical framework for
  transformer circuits. \emph{Transformer Circuits Thread}, Anthropic.
  https://transformer-circuits.pub/2021/framework/index.html
\item
  Engels, J., Riggs, L., \& Tegmark, M. (2025a). Decomposing the dark
  matter of sparse autoencoders. \emph{Transactions on Machine Learning
  Research (TMLR)}, April 2025. \emph{arXiv preprint arXiv:2410.14670}.
  https://arxiv.org/abs/2410.14670
\item
  Engels, J., Michaud, E. J., Liao, I., Gurnee, W., \& Tegmark, M.
  (2025b). Not all language model features are one-dimensionally linear.
  \emph{Proceedings of the International Conference on Learning
  Representations (ICLR 2025)}. \emph{arXiv preprint arXiv:2405.14860}.
  https://arxiv.org/abs/2405.14860
\item
  Geva, M., Schuster, R., Berant, J., \& Levy, O. (2021). Transformer
  feed-forward layers are key-value memories. \emph{Proceedings of the
  2021 Conference on Empirical Methods in Natural Language Processing
  (EMNLP 2021)}, 5484--5495. \emph{arXiv preprint arXiv:2012.14913}.
  https://arxiv.org/abs/2012.14913
\item
  Geva, M., Caciularu, A., Wang, K. R., \& Goldberg, Y. (2022).
  Transformer feed-forward layers build predictions by promoting
  concepts in the vocabulary space. \emph{Proceedings of the 2022
  Conference on Empirical Methods in Natural Language Processing (EMNLP
  2022)}, 30--45. \emph{arXiv preprint arXiv:2203.14680}.
  https://arxiv.org/abs/2203.14680
\item
  Goldowsky-Dill, N., MacLeod, C., Sato, L., \& Arora, A. (2023).
  Localizing model behavior with path patching. \emph{arXiv preprint
  arXiv:2304.05969}. https://arxiv.org/abs/2304.05969
\item
  Gurnee, W., Horsley, T., Guo, Z. C., Kheirkhah, T. R., Sun, Q.,
  Hathaway, W., Nanda, N., \& Bertsimas, D. (2024). Universal neurons in
  GPT2 language models. \emph{Transactions on Machine Learning Research
  (TMLR)}. \emph{arXiv preprint arXiv:2401.12181}.
  https://arxiv.org/abs/2401.12181
\item
  Gurnee, W., Ameisen, E., Kauvar, I., Tarng, J., Pearce, A., Olah, C.,
  \& Batson, J. (2026). When models manipulate manifolds: The geometry
  of a counting task. \emph{Transformer Circuits Thread}, Anthropic,
  October 2025 (arXiv archival 2026). \emph{arXiv preprint
  arXiv:2601.04480}. https://arxiv.org/abs/2601.04480
\item
  Henry, J. (2026, dataset). Rosetta Activations: Pre-extracted
  transformer residual stream activations for 46 language models across
  17 concepts (40 with full per-model analysis; 6 frontier models
  CAZ-only). HuggingFace.
  https://huggingface.co/datasets/james-ra-henry/Rosetta-Activations
  (DOI: https://doi.org/10.57967/hf/9725)
\item
  Henry, J. (2026, software). rosetta\_tools (v1.3.1). Zenodo.
  https://doi.org/10.5281/zenodo.20361433
\item
  Henry, J. (2026, software: Rosetta\_Analysis). Rosetta\_Analysis:
  Analysis, figure, and reproduction scripts for the Rosetta
  interpretability research program (v1.1.0). Zenodo.
  https://doi.org/10.5281/zenodo.21583139
\item
  Henry, J. (2026b). Geometric Evolution Maps: Extracting Stable Concept
  Probes from Transformer Residual Streams. \emph{arXiv preprint
  arXiv:2605.25848}. https://arxiv.org/abs/2605.25848
\item
  Henry, J. (2026c). Concept Encoding Strategies Across 28 Transformers:
  A Concept Allocation Zone and Geometric Evolution Map Evaluation.
  Zenodo. https://doi.org/10.5281/zenodo.22016917
\item
  Huh, M., Cheung, B., Wang, T., \& Isola, P. (2024). Position: The
  Platonic Representation Hypothesis. \emph{Proceedings of the 41st
  International Conference on Machine Learning (ICML 2024)}, PMLR 235,
  20617--20642. \emph{arXiv preprint arXiv:2405.07987}.
  https://arxiv.org/abs/2405.07987
\item
  Kim, B., Wattenberg, M., Gilmer, J., Cai, C., Wexler, J., Viégas, F.,
  \& Sayres, R. (2018). Interpretability beyond feature attribution:
  Quantitative testing with concept activation vectors (TCAV). In
  \emph{Proceedings of the 35th International Conference on Machine
  Learning (ICML 2018)}, 2668--2677. \emph{arXiv preprint
  arXiv:1711.11279}. https://arxiv.org/abs/1711.11279
\item
  Kornblith, S., Norouzi, M., Lee, H., \& Hinton, G. (2019). Similarity
  of neural network representations revisited. \emph{Proceedings of the
  International Conference on Machine Learning (ICML 2019)}, 3519--3529.
\item
  Mahalanobis, P. C. (1936). On the generalized distance in statistics.
  \emph{Proceedings of the National Institute of Sciences of India},
  2(1), 49--55.
\item
  Marks, S., \& Tegmark, M. (2024). The geometry of truth: Emergent
  linear structure in large language model representations of true/false
  datasets. \emph{Conference on Language Modeling (COLM 2024)}.
  \emph{arXiv preprint arXiv:2310.06824}.
  https://arxiv.org/abs/2310.06824
\item
  Meng, K., Bau, D., Andonian, A., \& Belinkov, Y. (2022). Locating and
  editing factual associations in GPT. \emph{Advances in Neural
  Information Processing Systems (NeurIPS 2022)}, 35, 17359--17372.
  \emph{arXiv preprint arXiv:2202.05262}.
  https://arxiv.org/abs/2202.05262
\item
  nostalgebraist. (2020). Interpreting GPT: The logit lens.
  \emph{LessWrong / AI Alignment Forum}, August 2020.
  https://www.lesswrong.com/posts/AcKRB8wDpdaN6v6ru/interpreting-gpt-the-logit-lens
\item
  Sofroniew, N., Kauvar, I., Saunders, W., Chen, R., Henighan, T.,
  Hydrie, S., Citro, C., Pearce, A., Tarng, J., Gurnee, W., Batson, J.,
  Zimmerman, S., Rivoire, K., Fish, K., Olah, C., \& Lindsey, J. (2026).
  Emotion concepts and their function in a large language model.
  \emph{Transformer Circuits Thread}, Anthropic. \emph{arXiv preprint
  arXiv:2604.07729}. https://arxiv.org/abs/2604.07729
\item
  Tenney, I., Das, D., \& Pavlick, E. (2019). BERT rediscovers the
  classical NLP pipeline. In \emph{Proceedings of the 57th Annual
  Meeting of the Association for Computational Linguistics (ACL 2019)},
  4593--4601. \emph{arXiv preprint arXiv:1905.05950}.
  https://arxiv.org/abs/1905.05950
\item
  Vig, J., Gehrmann, S., Belinkov, Y., Qian, S., Nevo, D., Singer, Y.,
  \& Shieber, S. (2020). Investigating gender bias in language models
  using causal mediation analysis. \emph{Advances in Neural Information
  Processing Systems (NeurIPS 2020)}, 33, 12388--12401.
\item
  Wang, K., Variengien, A., Conmy, A., Shlegeris, B., \& Steinhardt, J.
  (2023). Interpretability in the wild: A circuit for indirect object
  identification in GPT-2 small. \emph{Proceedings of the International
  Conference on Learning Representations (ICLR 2023)}. \emph{arXiv
  preprint arXiv:2211.00593}. https://arxiv.org/abs/2211.00593
\item
  Wollschläger, T., Elstner, J., Geisler, S., Cohen-Addad, V.,
  Günnemann, S., \& Gasteiger, J. (2025). The geometry of refusal in
  large language models: Concept cones and representational
  independence. \emph{Proceedings of Machine Learning Research (ICML
  2025)}, 267, 66945--66970. \emph{arXiv preprint arXiv:2502.17420}.
  https://arxiv.org/abs/2502.17420
\item
  Zhao, J., Huang, J., Wu, Z., Bau, D., \& Shi, W. (2025). LLMs encode
  harmfulness and refusal separately. \emph{arXiv preprint
  arXiv:2507.11878}. https://arxiv.org/abs/2507.11878
\item
  Zou, A., Phan, L., Chen, S., Campbell, J., Guo, P., Ren, R., Pan, A.,
  Yin, X., Mazeika, M., Dombrowski, A.-K., Goel, S., Li, N., Byun, M.
  J., Wang, Z., Mallen, A., Basart, S., Koyejo, S., Song, D.,
  Fredrikson, M., Kolter, J. Z., \& Hendrycks, D. (2023). Representation
  engineering: A top-down approach to AI transparency. \emph{arXiv
  preprint arXiv:2310.01405}. https://arxiv.org/abs/2310.01405
\end{itemize}

\end{document}